\documentclass[runningheads]{llncs}

\usepackage{eccv}

\usepackage{eccvabbrv}

\usepackage{graphicx}
\usepackage{booktabs}

\usepackage[accsupp]{axessibility}  

\usepackage{color, colortbl}
\definecolor{LightGray}{rgb}{0.88,0.88,0.88}
\usepackage{xcolor, soul}
\sethlcolor{LightGray}

\usepackage{hyperref}

\usepackage{orcidlink}

\usepackage{multirow}
\usepackage{tabularray}
\usepackage{amsmath,amssymb}
\usepackage{pifont}
\newcommand{\cmark}{\ding{51}}%
\newcommand{\xmark}{\ding{55}}%
\usepackage{graphicx}
\usepackage{float}

\begin{document}

\title{BAMM: Bidirectional Autoregressive Motion Model} 

\titlerunning{BAMM}

\author{Ekkasit Pinyoanuntapong\inst{1} \and
Muhammad Usama Saleem \inst{1} \and
Pu Wang \inst{1} \and
Minwoo Lee\inst{1} \and
Srijan Das\inst{1} \and
Chen Chen \inst{2}}

\authorrunning{E. Pinyoanuntapong et al.}

\institute{University of North Carolina at Charlotte \email{\{epinyoan,msaleem2, Pu.Wang, minwoo.lee, sdas24\}@uncc.edu} \and
University of Central Florida \email{chen.chen@crcv.ucf.edu}}

\maketitle

 \begin{figure*}[ht]
 \centering
  \includegraphics[width=1\textwidth]{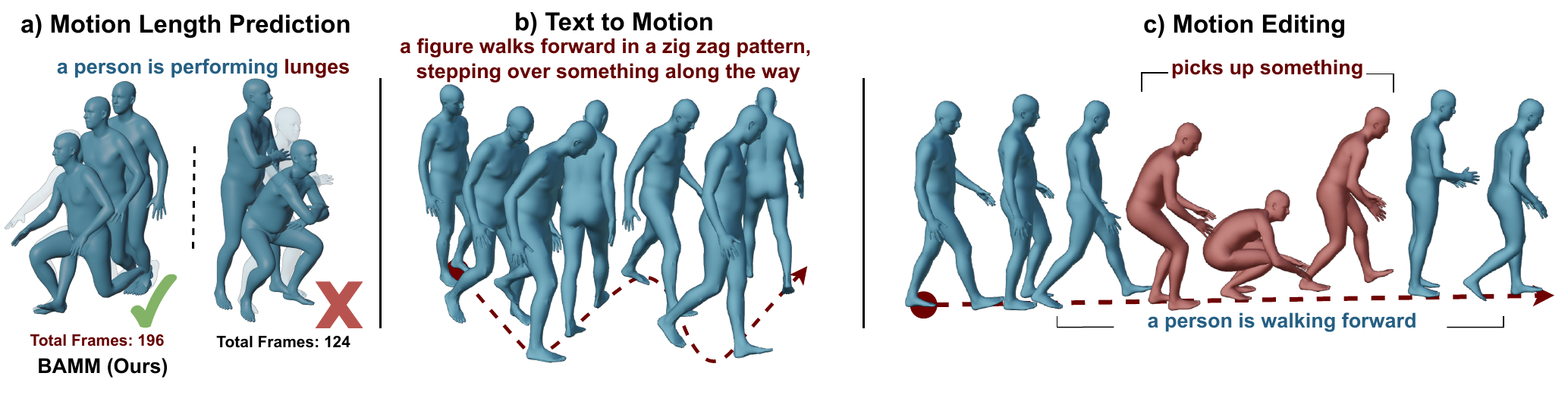}
  \vspace{-15pt}
  \caption{(a) Motion Length Prediction: Text-to-motion models often require specific input lengths, making them sensitive to motion generation. In contrast, BAMM automatically predicts the end of the motion, thus avoiding reliance on inaccurate motion length estimations. (b) High-quality Text-to-Motion: BAMM generates natural human movements precisely aligned with detailed textual descriptions. (c) Motion Editing: BAMM is capable of multiple editing tasks, such as inpainting (as demonstrated), outpainting, prefix prediction, suffix completion, and arbitrarily long motion sequence synthesis. }
  \label{fig:bamm_landing}
  \vspace{-25pt}
\end{figure*}

\begin{abstract}
  Generating human motion from text has been dominated by denoising motion models either through diffusion or generative masking process. However, these models face great limitations in usability by requiring prior knowledge of the motion length. Conversely, autoregressive motion models address this limitation by adaptively predicting motion endpoints, at the cost of degraded generation quality and editing capabilities. To address these challenges, we propose Bidirectional Autoregressive Motion Model (BAMM), a novel text-to-motion generation framework. BAMM consists of two key components: (1) a motion tokenizer that transforms 3D human motion into discrete tokens in latent space, and (2) a masked self-attention transformer that autoregressively predicts randomly masked tokens via a hybrid attention masking strategy. By unifying generative masked modeling and autoregressive modeling, BAMM captures rich and bidirectional dependencies among motion tokens, while learning the probabilistic mapping from textual inputs to motion outputs with dynamically-adjusted motion sequence length.  This feature enables BAMM to simultaneously achieving high-quality motion generation with enhanced usability and built-in motion editability. Extensive experiments on HumanML3D and KIT-ML datasets demonstrate that BAMM surpasses current state-of-the-art methods in both qualitative and quantitative measures. Our project page is available at \url{https://exitudio.github.io/BAMM-page}
  
\end{abstract}

\section{Introduction}
\label{sec:intro}

Text-to-motion generation is a promising interdisciplinary field that uses natural language to generate 3D human movements. This emerging field holds vast potential to transform animation, gaming, filming, and VR/AR/MR domains by enabling easy and intuitive creation of 3D assets through user-friendly textual inputs. However, bridging the semantic gap between textual descriptions and intricate motion sequences presents a significant challenge. To address this challenge, recent efforts have been focusing on two methods: (1) conditional denoising motion model and (2) conditional autoregressive motion model. Both methods can greatly improve motion generation quality by learning the probabilistic distribution of motion sequences, conditioned on the textural descriptors. However, both methods face fundamental limitations. 



Conditional denoising motion models are trained to restore corrupted motion sequences to their original state, guided by textual prompts. These models operate through two primary mechanisms: diffusion and generative masking. Diffusion models apply structured Gaussian noise to the original motion data for corruption  \cite{MDM, MotionDiffuse, FLAME, Fg-T2M}, whereas generative masked models \cite{MMM, MoMask} corrupt motion sequence by substituting selected motion tokens with \texttt{[MASK]} tokens. The denoising process, followed by motion corruption procedure, considers motion tokens from both directions, effectively capturing the intricate dependencies among tokens. This leads to enhanced motion generation quality. Furthermore, these models inherently ease the motion editing tasks. By selectively corrupting and recovering motion tokens in areas needing modifications, they ensure seamless transitions between edited and unedited segments.


While denoising motion models excel in generation quality and editability, a significant limitation of these models is their usability because these models depend on prior knowledge of motion length for each text prompt, a requirement that proves impractical in real-world scenarios. Utilizing incorrect motion lengths can result in a significant decline in generation quality. To address this fundamental limitation, conditional motion autoregressive models emerge as a solution, capable of simultaneously predicting the sequence length and content of generated motions. Inspired by large language models like GPT \cite{GPT}, motion autoregressive models sequentially predict one motion token at a time from left to right until the \texttt{[END]} token is predicted, guided by the textual description \cite{T2M-GPT,AttT2M, MotionGPT}. As a result, the generated motions are not only well aligned with the text inputs but also appropriately scaled in duration. However, the sequential token decoding of autoregressive models cannot fully capture
the dependencies between the motion tokens, 
potentially compromising generation quality, and complicating the editing process, because edited parts need to be conditioned on the
unedited parts from both directions to ensure overall continuity and coherence.

As previously highlighted, existing text-driven motion generation models encounter a comprise among usability, quality of generation, and editability. To address this challenge, we propose Bidirectional Autoregressive Motion Model (BAMM), a novel text-to-motion generation framework. It consists of two pivotal components: a motion tokenizer and a conditional masked self-attention transformer. In a two-stage training paradigm, the motion tokenizer performs initial training based on Vector Quantized Variational Autoencoders (VQ-VAE) \cite{vqvae}. This tokenizer encodes the raw motion sequence into discrete motion tokens within the latent space leveraging learned codebooks.  In the subsequent phase, motion tokens are randomly masked out. A conditional masked self-attention transformer is then trained to autoregressively predict these masked tokens, adopting the causal attention masking strategy. This
strategy is largely departing from the traditional \texttt{[MASK]} token replacement approach in generative masked models. In particular, it does not substitute the input motion tokens with \texttt{[MASK]} tokens. Instead, it adjusts the attention score matrix according to both unidirectional and bidirectional causal masks. The unidirectional causal mask enables adaptive prediction of \texttt{[END]} token based on text prompts, while bidirectional causal mask forces the model to predict the next motion token not only based on past tokens but also conditioned on future unmasked tokens. This facilitates bidirectional autoregressive training for enhanced predictive capabilities. 
\begin{table}
\centering
\caption{Comparison of quality and capability of generation on text-to-motion to state-of-the-art models on the largest text-to-motion dataset \cite{t2m}. `\textcolor{red}{\cmark}' means capability while `\xmark' is not. "Predict Length" denotes the ability to generate motion without prior knowledge of motion length. "Input Length" refers to the ability to take input length as a constraint, while "Edit" indicates motion editability. Since MMM and MoMask require ground-truth motion length as input, we use predicted motion length from pretrained length estimator by \cite{t2m}. The lowest FID score means the best overall quality of the generated motion, ensuring that its
authenticity and naturalness is very close to the ground-truth human movements.  High R-precision and
low MM-dist means accurate alignment between
the generated motion and the text prompts.
} 
\label{tab:speed_all_models}
\scalebox{.6}{
\begin{tblr}{
  cell{1}{1-7} = {c},
  cell{2}{1-7} = {c},
  cell{3}{1-7} = {c},
  cell{4}{1-7} = {c},
  cell{5}{1-7} = {c},
  cell{6}{1-7} = {c},
  hline{1-2,6-7} = {-}{},
}
Methods     & {Top-1 $\uparrow$} & FID $\downarrow$ & MM-Dist $\downarrow$ & Predict Length & Input Length & Edit \\
T2M-GPT     & 0.491 & 0.116 & 3.118   & \textcolor{red}{\cmark} & \xmark &    \xmark  \\
AttT2M      & 0.499 & 0.112 & 3.038   & \textcolor{red}{\cmark} & \xmark &    \xmark  \\
MMM         & 0.504 & 0.080 & 2.998   & \xmark & \textcolor{red}{\cmark} & \textcolor{red}{\cmark}\\
MoMask      & 0.522 & 0.090 & 2.945   & \xmark & \textcolor{red}{\cmark} & \textcolor{red}{\cmark} \\
BAMM (Ours) & $\textcolor{red}{\mathbf{0.525}}$ & $\textcolor{red}{\mathbf{0.055}}$ & $\textcolor{red}{\mathbf{2.919}}$ & \textcolor{red}{\cmark} & \textcolor{red}{\cmark} & \textcolor{red}{\cmark}
\end{tblr}
}
\end{table}

By unifying masked and autoregressive prediction during training, BAMM captures rich and bidirectional dependencies among motion tokens, while learning a direct probabilistic mapping from textual inputs to motion outputs with dynamically-adjusted motion sequence length. Leveraging such unique feature, we propose cascaded motion generation during inference, where BAMM first leverages unidirectional autoregressive decoding to implicitly predict motion sequence length and generate coarse-grained motion sequence. Such motion sequence is then refined by masking and regenerating a portion of motion tokens in a bidirectional autoregressive fashion. This feature allows BAMM to achieve high-quality motion generation with high usability. Moreover, BAMM naturally supports zero-shot motion editing without specially being trained for such task. By treating the masked motion tokens as the contents that need editing, BAMM can predict the masked tokens based on the surrounding context and the text description. Our main contributions can be summarized as follows.

\begin{itemize}
    \item We introduce the bidirectional autoregressive motion model, which is a novel text-to-motion generation framework. It effectively harnesses the complementary benefits of denoising and autoregressive models, thus simultaneously achieving high-quality motion generation with enhanced usability and  innate motion editability, as showcased in Fig. \ref{fig:bamm_landing} and Table \ref{tab:speed_all_models}.

    \item We demonstrate that our model outperforms current state-of-the-art methods qualitatively and quantitatively on two standard text-to-motion generation datasets, HumanML3D \cite{t2m} and KIT-ML \cite{KIT}.
    
    \item We showcase that our model supports  a variety of motion editing tasks in the zero-shot manner without specially training for these tasks, including motion inpainting, outpainting, prefix prediction, suffix completion, and long sequence generation.
\end{itemize}
\section{Related Work}
\label{sec:related_work}

\textbf{Motion Synthesis with Latent Space Alignment.} Early text-to-motion approaches typically adopt a two-stage scheme: learning separate latent representations for text and motion sequences, followed by latent space alignment using distance losses like cosine similarity or KL divergence  \cite{Language2Pose, TEMOS,t2m,MotionCLIP,TMR,CrossModalRF}. For example, Language2Pose \cite{Language2Pose} aimed to establish a shared latent space for both language descriptions and motion sequences. MotionCLIP \cite{MotionCLIP} simplified this concept by incorporating stylization and diversity techniques, leveraging the pre-trained text-image latent space of the CLIP model \cite{CLIP}.  However, this strategy inherently struggles with generating high-fidelity motions due to the difficulty of perfectly aligning these inherently disparate latent spaces.


\textbf{Conditional Denoising Motion Model.} 
Inspired by the success of denoising diffusion models (DDMs) \cite{DDIM, DDPM} in text-to-image and text-to-video generation \cite{GLIDE,Imagen, Make-A-Video, ImagenVideo}, diffusion models have been applied for text-to-motion generation. MDM  \cite{MDM}, MotionDiffuse \cite{MotionDiffuse}, MLD \cite{MLD}, and FRAME \cite{FLAME} are the representative examples. Meanwhile, BERT-type masked generative models have shown their success in both text generation tasks, e.g., Q\&A and language translations \cite{mask-predict,GlancingTF,BERT}, and text-to-image synthesis \cite{M6UFCUM,UFCBERTUM,CogView2FA,MaskGIT, Phenaki, Parti, Muse}. Following this trend, MMM \cite{MMM} and MoMask \cite{MoMask} are recent attempts to propose conditional masked motion modeling to enable high-fidelity motion synthesis that is precisely aligned with text prompts. Both diffusion and generative masked modeling share the same principle: ``denoising'' corrupted data \cite{DiffusMask}. Meanwhile, they also face the same usability limitation: they require prior knowledge of the motion sequence length for the denoising process. This length information is obtained either by intuitive guess from users or via a separately pre-trained predictor that predicts the motion length distribution based on the input text prompts \cite{t2m}. Both approaches, however, lead to significant motion quality degradation as demonstrated in the experiments section. This is because the distributions of motion length and motion contents are inherently coupled, which need to be jointly aligned with the textual context.

\textbf{Conditional Autoregressive Motion Model.} The autoregressive motion models, such as  T2M-GPT \cite{T2M-GPT}, AttT2M \cite{AttT2M}, and MotionGPT \cite{MotionGPT}, can effectively address the usability challenge faced by denoising models because they follow the GPT-type training and inference \cite{GPT} to implicitly predict motion sequence length by generating the \texttt{[END]} token conditioned on both previously generated motion tokens and text inputs. However, the limitation of these models lies in their use of causal attention for unidirectional and sequential motion token prediction. This practice not only hinders model's motion editability but also jeopardizes motion generation quality. To address this limitation, we propose the first bidirectional autoregressive modeling approach for human motion generation, which draws inspiration from the self-attention masking employed by large language model pretraining \cite{Vaswani2017AttentionIA, Du2021GLMGL}.

\section{Method}
\label{sec:method}
\vspace{-5mm}
\begin{figure*}[ht]
\begin{center}
\centerline{\includegraphics[width=1\textwidth]{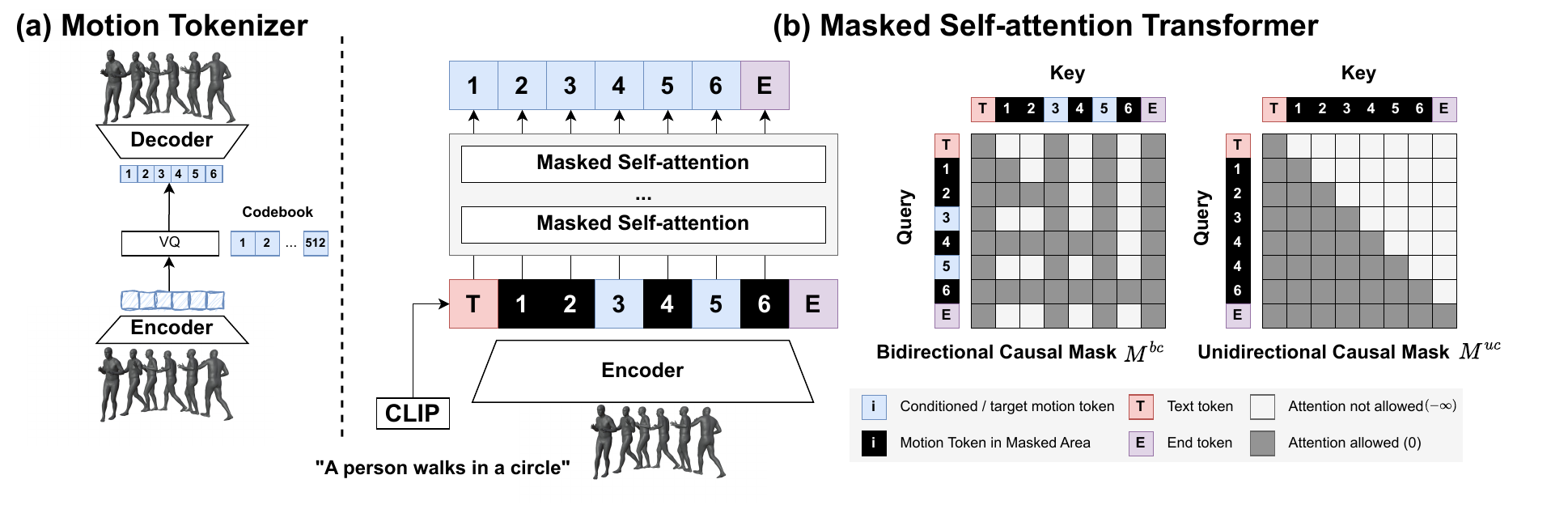}
}
\end{center}
\label{fig:training}
\vspace{-30pt}
\caption{Overall architecture of BAMM. {\bf (a) Motion Tokenizer} encodes the raw motion sequence into discrete motion tokens according to a learned codebook. {\bf (b) Masked Self-attention Transformer} learns to sequentially predict next tokens conditioned on text embedding from CLIP model and future unmasked tokens. Masked self-attention mechanism unifies autoregressive model and generative masked motion via bidirectional and unidirectional causal masks.}
\label{fig:training}
\vspace{-10pt}
\end{figure*}
Our objective is to create a text-to-motion synthesis framework that simultaneously achieves high-quality motion generation with enhanced usability and innate motion editability. Towards this goal, our framework, as illustrated in Fig. \ref{fig:training}, consists of two key components: motion tokenizer that compresses and converts raw 3D human motion into a sequence of discrete motion tokens in the latent space (Section \ref{sec:Motion_Tokenizer}) and conditional masked self-attention transformer that leverages unidirectional and bidirectional casual masks to integrate autoregressive model and generative masked model into a unified framework (Section \ref{sec:Masked_Self-attention_Transformer}).  The training procedure follows a hybrid attention masking strategy, where the two causal masks are applied randomly and the model is forced to reconstruct the motion sequence under both cases (Section \ref{sec:training}). The cascaded motion decoding is introduced for motion generation during the inference phase. It uses unidirectional autoregressive decoding to jointly predict motion sequence length and its contents, which are refined via bidirectional autoregressive decoding (Section \ref{sec:inference}).




\subsection{Motion Tokenizer} 
\label{sec:Motion_Tokenizer}
The objective of this stage is to learn the discrete representation of motion by quantizing the embedding $z$ from the output of the encoder into codebook $\mathcal{C}$. We first pretrain a motion tokenizer based on VQ-VAE \cite{vqvae}. In particular, given a motion sequence $\mathcal{M} = [m_1, m_2, m_3, ..., m_\tau]$ where $m \in \mathbb{R}^{D}$, $\tau$ is the total frames of motion, and $D$ is the dimension of the 3D pose in each frame, encoder is used to encode motion $\mathcal{M}$ to the latent embedding $z \in \mathbb{R}^{t \times d}$ with a downsampling rate of $\tau/t$. The embedding $z$ is quantized into codes $c \in \mathcal{C}$. Codebook $\mathcal{C} = \{\gamma_k\}^K_{k=1}$ contains $K$ number of codes. The nearest Euclidean distance between the embedding $\mathbf{z}$ and the code of vector is computed by $\hat{z_i}=\operatorname{argmin}_j\left\|\mathbf{z}-\mathbf{\mathcal{C}}_j\right\|_2^2$. The loss function is defined as
\begin{equation}\label{eq:vector-quantization}
L_{V Q}=\|\operatorname{sg}(\mathbf{z})-\mathbf{e}\|_2^2+\beta\|\mathbf{z}-\operatorname{sg}(\mathbf{e})\|_2^2,
\end{equation}
where $\operatorname{sg}(\cdot ) $ is the stop-gradient operator, $\beta$ refers the hyper-parameter for commitment loss. The loss function is optimized via a straight-through gradient estimator. We apply exponential moving average for codebooks update and codebook reset by following \cite{T2M-GPT} \cite{MoMask} \cite{MMM}.

\subsection{Conditional Masked Self-attention Transformer}
\label{sec:Masked_Self-attention_Transformer}
Our model employs a standard multi-layer transformer, whose inputs are the concatenation of the motion tokens $x_{1:t}$ from the tokenizer with $t$ as the sequence length, the text embedding $x_0$ from the pre-trained CLIP model \cite{CLIP}, and the \texttt{[END]} token $x_{t+1}$ that serves as the indicator of the motion’s endpoint. The input tokens $x_{0:t+1}$ are masked out strategically. Different from generative masked models, we do not replace the input tokens with \texttt{[MASK]} ones. Instead, we adopt causal attention mask $M$ as shown in Fig. \ref{fig:training} (b) to specify the attention relations among the input tokens. In particular, the token in the masked areas, indicated by $\blacksquare$, can be attended to itself, all the tokens on its left, and the unmasked tokens on its right. The unmasked token can be attended to by other unmasked tokens in both directions. In particular, we employ two causal masks: the unidirectional one, where only text token is unmasked and all other tokens are in the masked areas, and the bidirectional one, where text and \texttt{[END]} tokens are unmasked, while a random number of motion tokens are put into the masked areas. Consequently, Masked Self-attention Transformer retains the causal aspect of autoregressive motion generation while also being capable of conditioning on future tokens. The output of  masked self-attention is as follows:

\begin{equation}\label{eq:attention_w_mask}
Attention =\operatorname{Softmax}\left(\frac{Q K^T}{\sqrt{d_k}}  + \text { M } \right ) \cdot V
\end{equation}
where Q, K, and V, indicate queries, keys, and values respectively while $d_k$ represents the dimension of queries and keys. The self-attention mask $M \in \mathbb{R}^{(t + 1) \times (t +1)}$ is assigned to zero in the positions where attention is allowed, and to negative infinity otherwise. Adding negative infinity forces attention score to be zero after $Softmax(\cdot)$ operation. Therefore, {\it bidirectional causal mask} $M^{bc}$ can be written as
\begin{equation}
\label{eq:masked_attention}
M_{ij} = \left\{
    \begin{array}{ll}
        0, & \text{where } (i \geq j \land i \notin U) \vee (j \in U)  \\
        -\infty, & \text{otherwise}
    \end{array}
\right.
\end{equation}
where $i, j \in [0, 1, 2, \ldots, t+1]$ is the index of query $Q$ and key $K$. $U = [u_0, u_1, ...]$ contains the indices of unmasked tokens. The {\it unidirectional causal mask} $M^{uc}$ is a special case of the bidirectional one when $U = \emptyset$.

\subsection{Training: Hybrid Attention Masking}
\label{sec:training}
Given a discrete representation of the motion sequence $x_{1:t}$, our model is trained to
reconstruct the motion sequence, conditioned on the text token $x_0$ under both unidirectional and bidirectional causal masking strategies, i.e., $M^{uc}$ and  $M^{bc}$. The reconstruction probability of each motion token under each masking case is $p_{\theta}(x_i \mid M^{uc})$ and $p_{\theta}(x_i \mid M^{bc})$, respectively.  The training objective is to minimize the negative log-likelihood of the motion sequence prediction 

\begin{equation}
\small
\mathcal{L}_{\text {hybrid}}=-\underset{\mathbf{X} \in p(\mathbf{X})}{\mathbb{E}}\left[\lambda\sum_{\forall i \in[1, t]} \log p_{\theta}(x_i \mid M^{uc}) + (1 - \lambda)\sum_{\forall i \in[1, t]} \log p_{\theta}(x_i \mid M^{bc})\right].
\end{equation}
where $\lambda$ is the probability of selecting unidirectional causal mask. Through experiments, we found $\lambda = 0.5$ yields best performance. In addition, when bidirectional causal mask is selected, we randomly put $50\% - 100\%$ motion tokens in the masked areas.

\subsection{Inference: Cascaded Motion Decoding}
\label{sec:inference}

\begin{figure*}[ht]
\begin{center}
\centerline{\includegraphics[width=.8\textwidth]{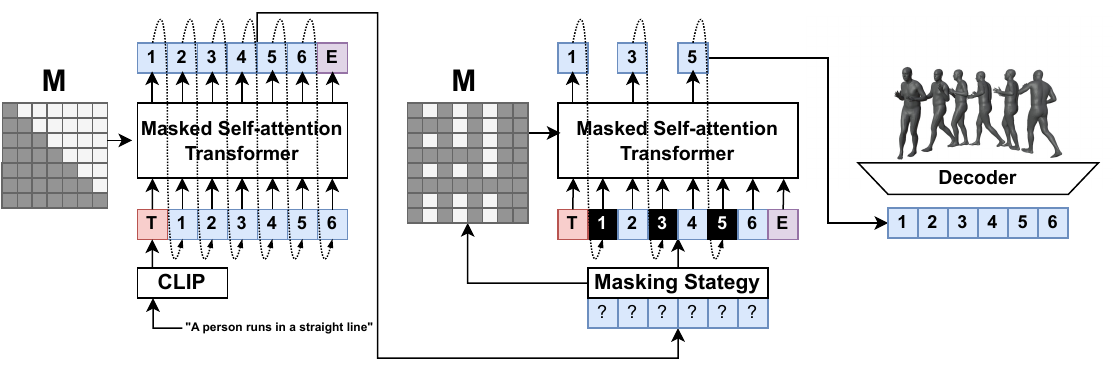}
}
\end{center}
\vspace{-30pt}
\caption{Inference: Dual-iteration Cascaded Motion Decoding. In the first iteration, 
autoregressive decoding is applied by adopting unidirectional causal mask  to generate coarse-grained motion and predict motion sequence length. In the second iteration,  bidirectional autoregressive decoding is performed via bidirectional causal mask to removing and repredicting low-confidence motion tokens autoregressively. }
\label{fig:inference}
\vspace{-10pt}
\end{figure*}
To generate motion sequence during inference phase, dual-iteration cascaded decoding is introduced. In the first iteration, autoregressive decoding is applied,  where the motion tokens are sequentially and stochastically sampled according to unidirectional prediction distribution $p_{\theta}(x_i \mid M^{uc})$, concluding upon predicting the \texttt{[END]} token.  Since autoregressive decoding could accumulate prediction errors from the previously generated tokens, the generated motion sequence is refined in the second iteration by masking out a subset of motion tokens and then resampling these masked tokens according to the bidirectional prediction distribution $p_{\theta}(x_i \mid M^{bc})$. In this refinement iteration, \texttt{[END]} is positioned where it was predicted in the first iteration. The model can attend to unmasked tokens in all directions to re-predict low-confidence tokens based on rich surrounding context. The choice of masking strategies has an impact on the refinement gain, which is evaluated in Section \ref{tab:ablation}.

{\bf Hybrid Classifier-free Guidance.} At training time, we randomly drop textual tokens to teach the model to generate motion unconditionally. During inference, we apply classifier-free guidance (CFG)  \cite{cfg} for cascaded motion decoding.  In particular, we generate the final motion sequence by a linear combination of the conditioned logits $\ell_c$ and unconditional logits $\ell_u$ with guidance scale $s$ as
\begin{equation}
\label{eq:cfg}
\ell_g = (1+s)\cdot \ell_c - s \cdot \ell_u.
\end{equation}
We apply different CFG scale $s$ for each iteration during cascaded decoding. The effectiveness of CFG in each iteration is evaluated in Section \ref{sec:ablation}.

\begin{figure*}[ht]
\begin{center}
\scalebox{1}{
\centerline{\includegraphics[width=1\textwidth]{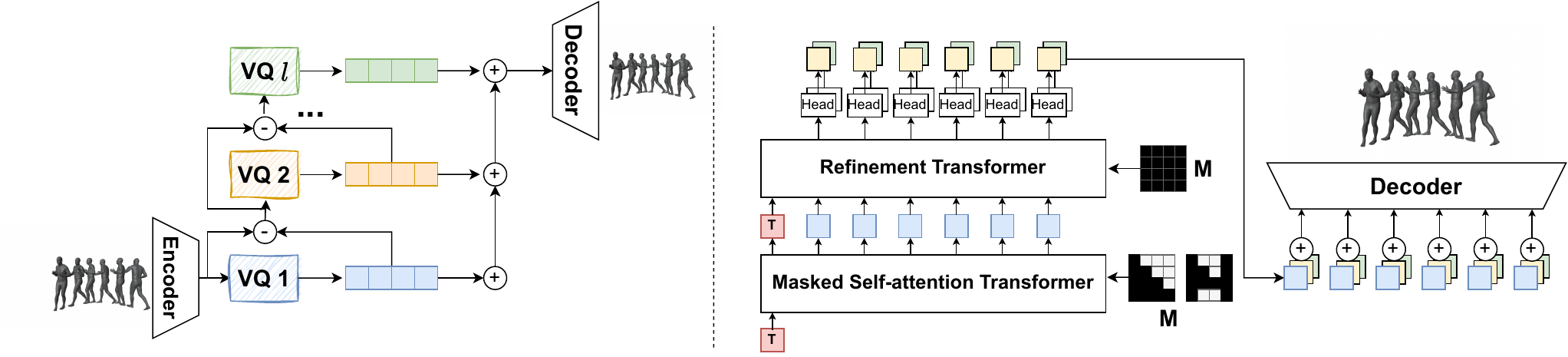}
}}
\end{center}
\vspace{-10pt}
\caption{Residual Motion Refinement. The residual vector quantization encodes the raw motion sequence into multiple token sequences in different colors (left). The base token sequence from the first vector quantizer is generated via cascaded decoding by masked self-attention transformer. The base token sequence is used as the input of the refinement transformer to predict the residual token sequences from other quantizers. The combined sequences are fed into tokenizer's decoder for motion generation. The refinement transformer shares the same architecture as the masked self-attention transformer with a full attention mask(right).  }
\label{fig:rvq}
\end{figure*}

{\bf Residual Motion Refinement.} To further enhance motion generation quality, the motion sequence yielded from cascaded decoding can be refined by another refinement transformer based on residual vector quantization (RVQ) \cite{Zeghidour2021SoundStreamAE}. By utilizing RVQ, the raw motion sequence is encoded into multiple token sequences in the latent space. Each token sequence is generated by a separate quantizer and each
quantizer encodes the quantization error left by the previous quantizer. As a result, the token sequence from the first quantizer encodes the most of information of the original motion sequence. The rest of the token sequences from other quantizers only encode quantization errors. Through RVQ, information loss can be minimized during the embedding quantization process. As a result, our masked self-attention transformer is used to generate the most informative token sequence from the first quantizer. Using this sequence as input, another refinement transformer is trained to predict the remaining token sequences, which are merged into a single token sequence for final motion decoding. This RVQ-based refinement has been adopted by audio generation models \cite{Du2021GLMGL,Wang2023NeuralCL,Borsos2023SoundStormEP} and recently demonstrates its benefits in motion generation task \cite{MoMask}.  

\subsection{Motion Editability}
The autoregressive approach naturally lacks the ability for temporal motion editing as it cannot leverage future motion tokens. In contrast, BAMM enables temporal motion editing through its bidirectional causal mask, which allows the model to access information from all directions of the conditioned tokens. Consequently, temporal motion editing becomes achievable by merely applying masked attention to the positions that require medications. We illustrate various editing tasks, i.e. motion inpainting (in-betweening), outpainting, prefix prediction, and suffix completion qualitatively and quantitively in Fig. \ref{fig:editing_tasks} and Table \ref{tab:edit}. BAMM also allows us to generate the arbitrarily long motion sequence according to a sequence of text prompts, which is showcased in \textbf{Supplementary Material}.
\section{Experiments}


In this section, we provide a comprehensive evaluation along with empirical results for our proposed motion generation model, BAMM. In Section \ref{subsec:comparision_sota}, we adhere to standard evaluation protocols on two datasets, demonstrating that our model outperforms current state-of-the-art methods both quantitatively and qualitatively. Furthermore, in Section \ref{sec:predlen_edit}, we showcase our predictive length and editing capabilities in comparison to state-of-the-art methods. The results reveal the robust performance of our model, displaying its effectiveness even in cases without prior information about motion length. 

\textbf{Datasets.}  We evaluated our model using the evaluation protocol proposed by \cite{t2m} for text-driven motion generation with two datasets. {\bf KIT-ML} dataset contains 3,911 motion sequences, each with one to four textual annotations, amounting to a total of 6,278 annotations. These motion sequences are derived from the KIT and CMU motion \cite{cmu-mocap} databases and have been adjusted to 12.5 FPS. The dataset is divided into training, validation, and testing sets, with proportions of 80\%, 5\%, and 15\% respectively. {\bf HumanML3D} dataset includes a wide variety of human activities, such as exercise and dancing, and consists of 14,616 motion sequences paired with 44,970 textual descriptions. These descriptions come from a vocabulary of 5,371 unique words. The motion sequences, sourced from AMASS \cite{AMASS} and HumanAct12 \cite{HumanAct12}, have been standardized to 20 FPS and are limited to a maximum duration of 10 seconds, with the actual lengths varying between 2 to 10 seconds. Each sequence is accompanied by at least three descriptive annotations, averaging 12 words in length.



\textbf{Evaluation Metrics.} We adopt the standard evaluation framework from T2M \cite{t2m}, employing pre-trained models that encode text and motion information to evaluate text and motion tokens in the embedding space. R-precision (Top-1, 2, 3 accuracy) measures how well the generated motions align with the text prompts, while Multimodal Distance (MM-Dist) quantifies the distance between generated and ground-truth motions in a shared feature space. Frechet Inception Distance (FID) assesses the statistical similarity between the feature distributions of generated and real motions. Additionally, we evaluate diversity (the average Euclidean distance between random motion pairs) and multimodality (the average variance across Euclidean distances between generated motion pairs for a single prompt) to capture the range of possible motion interpretations and their consistency with the text description.


\subsection{Comparison to State-of-the-art Approaches}
\label{subsec:comparision_sota}

\begin{table*}[ht]
\centering
\caption{\textbf{Comparison of text-conditional motion synthesis on HumanML3D \cite{t2m} test set.} We repeat the evaluation 20 times for each metric and report the average with 95$\%$ confidence interval. \textcolor{red}{Red} and \textcolor{blue}{Blue} indicate the best and the second best result. Methods with \hl{gray highlight $^{\S}$} report motion generation results using the ground-truth motion length.}
\vspace{-7pt}
\scalebox{0.7}{
\begin{tabular}{lccccccc} 
\hline
\multirow{2}{*}{Methods} & \multicolumn{3}{c}{R-Precision $\uparrow$} & \multirow{2}{*}{FID $\downarrow$} & \multirow{2}{*}{MM-Dist $\downarrow$} & \multirow{2}{*}{Diversity $\uparrow$} & \multirow{2}{*}{MModality $\uparrow$}        \\ 
\cline{2-4}
 & Top-1 $\uparrow$ & Top-2 $\uparrow$ & Top-3 $\uparrow$ & & & & \\ 
\toprule

Hier \cite{hier} & $0.301^{\pm .002}$ & $0.425^{\pm .002}$ & $0.552^{\pm .004}$ & $6.523^{\pm .024}$ & $5.012^{\pm .018}$ & $8.332^{\pm .042}$ & - \\
\rowcolor{LightGray}
TEMOS$^{\S}$ \cite{TEMOS} & $0.424^{\pm .002}$ & $0.612^{\pm .002}$ & $0.722^{\pm .002}$ & $3.734^{\pm .028}$ & $3.703^{\pm .008}$ & $8.973^{\pm .071}$ & $0.368^{\pm .018}$ \\
TM2T \cite{TM2T} & $0.424^{\pm .003}$ & $0.618^{\pm .003}$ & $0.729^{\pm .002}$ & $1.501^{\pm .017}$ & $3.467^{\pm .011}$ & $8.589^{\pm .076}$ & $2.424^{\pm .093}$ \\
T2M \cite{t2m} & $0.455^{\pm .003}$ & $0.636^{\pm .003}$ & $0.736^{\pm .002}$ & $1.087^{\pm .021}$ & $3.347^{\pm .008}$ & $9.175^{\pm .083}$ & $2.219^{\pm .074}$ \\
\rowcolor{LightGray}
MDM$^{\S}$ \cite{MDM} & $0.320^{\pm .005}$ & $0.498^{\pm .004}$ & $0.611^{\pm .007}$ & $0.544^{\pm .044}$ & $5.566^{\pm .027}$ & $9.559^{\pm .086}$ & $\mathbf{\textcolor{blue}{2.799^{\pm .072}}}$ \\
\rowcolor{LightGray}
MotionDiffuse$^{\S}$ \cite{MotionDiffuse} & $0.491^{\pm .001}$ & $0.681^{\pm .001}$ & $0.782^{\pm .001}$ & $0.630^{\pm .001}$ & $3.113^{\pm .001}$ & $9.410^{\pm .049}$ & $1.553^{\pm .042}$ \\
\rowcolor{LightGray}
MLD$^{\S}$ \cite{MLD} & $0.481^{\pm .003}$ & $0.673^{\pm .003}$ & $0.772^{\pm .002}$ & $0.473^{\pm .013}$ & $3.196^{\pm .010}$ & $9.724^{\pm .082}$ & $2.413^{\pm .079}$ \\
\rowcolor{LightGray}
Fg-T2M$^{\S}$ \cite{Fg-T2M} & $0.492^{\pm .002}$ & $0.683^{\pm .003}$ & $0.783^{\pm .002}$ & $0.243^{\pm .019}$ & $3.109^{\pm .007}$ & $9.278^{\pm .072}$ & $1.614^{\pm .049}$ \\
\rowcolor{LightGray}
M2DM$^{\S}$ \cite{M2DM} & $0.497^{\pm .003}$ & $0.682^{\pm .002}$ & $0.763^{\pm .003}$ & $0.352^{\pm .005}$ & $3.134^{\pm .010}$ & $\mathbf{\textcolor{red}{9.926^{\pm .073}}}$ & $\textcolor{red}{\mathbf{3.587^{\pm .072}}}$ \\
T2M-GPT \cite{T2M-GPT} & $0.491^{\pm .003}$ & $0.680^{\pm .003}$ & $0.775^{\pm .002}$ & $0.116^{\pm .004}$ & $3.118^{\pm .011}$ & $\mathbf{\textcolor{blue}{9.761^{\pm .081}}}$ & $1.856^{\pm .011}$ \\
AttT2M \cite{AttT2M} & $0.499^{\pm .003}$ & $0.690^{\pm .002}$ & $0.786^{\pm .002}$ & $0.112^{\pm .006}$ & $3.038^{\pm .007}$ & $9.700^{\pm .090}$ & $2.452^{\pm .051}$ \\
\rowcolor{LightGray}
MMM$^{\S}$ \cite{MMM} & $0.515^{\pm .002}$ & $0.708^{\pm .002}$ & $0.804^{\pm .002}$ & $0.089^{\pm .005}$ & $\mathbf{\textcolor{blue}{2.926^{\pm .007}}}$ & $9.577^{\pm .050}$ & $1.226^{\pm .035}$ \\
\rowcolor{LightGray}
MoMask$^{\S}$ \cite{MoMask} & $0.521^{\pm .002}$ & $0.713^{\pm .003}$ & $0.807^{\pm .002}$ & $\mathbf{\textcolor{red}{0.045^{\pm .002}}}$ & $2.958^{\pm .008}$ & $-$ & $1.241^{\pm .040}$ \\

\toprule
BAMM (ours) & $\mathbf{\textcolor{red}{0.525^{\pm .002}}}$ & $\mathbf{\textcolor{red}{0.720^{\pm .003}}}$ & $\mathbf{\textcolor{red}{0.814^{\pm .003}}}$ & $\mathbf{\textcolor{blue}{0.055^{\pm .002}}}$ & $\mathbf{\textcolor{red}{2.919^{\pm ..008}}}$ & $9.717^{\pm .089}$ & $1.687^{\pm .051}$ \\
\rowcolor{LightGray}
BAMM$^{\S}$ (ours) & $\mathbf{\textcolor{blue}{0.522^{\pm .003}}}$ & $\mathbf{\textcolor{blue}{0.715^{\pm .003}}}$ & $\mathbf{\textcolor{blue}{0.808^{\pm .003}}}$ & $\mathbf{\textcolor{blue}{0.055^{\pm .002}}}$ & $2.936^{\pm .077}$ & $9.636^{\pm .009}$ & $1.732^{\pm .055}$ \\
\bottomrule
\end{tabular}
}
\label{tab:humanml3d}
\end{table*}

\begin{table*}[h!]
\centering
\caption{\textbf{Comparison of text-conditional motion synthesis on KIT-ML \cite{KIT} test set.} We repeat the evaluation 20 times for each metric and report the average with 95$\%$ confidence interval. \textcolor{red}{Red} and \textcolor{blue}{Blue} indicate the best and the second best result. Methods with \hl{gray highlight $^{\S}$} report ground-truth motion length for generation.}
\vspace{-7pt}
\scalebox{0.7}{
\begin{tabular}{lccccccc} 
\hline
\multirow{2}{*}{Methods} & \multicolumn{3}{c}{R-Precision $\uparrow$} & \multirow{2}{*}{FID $\downarrow$} & \multirow{2}{*}{MM-Dist $\downarrow$} & \multirow{2}{*}{Diversity $\uparrow$} & \multirow{2}{*}{MModality $\uparrow$}        \\ 
\cline{2-4}
& Top-1 $\uparrow$& Top-2 $\uparrow$& Top-3 $\uparrow$& & & & \\ 
\toprule
Hier \cite{hier} & $0.255^{\pm .006}$ & $0.432^{\pm .007}$ & $0.531^{\pm .007}$ & $5.203^{\pm .107}$ & $4.986^{\pm .027}$ & $9.563^{\pm .072}$ & - \\
\rowcolor{LightGray}
TEMOS$^{\S}$ \cite{TEMOS} & $0.353^{\pm .006}$ & $0.561^{\pm .007}$ & $0.687^{\pm .005}$ & $3.717^{\pm .051}$ & $3.417^{\pm .019}$ & $10.84^{\pm .100}$ & $0.532^{\pm .034}$ \\
TM2T \cite{TM2T} & $0.280^{\pm .005}$ & $0.463^{\pm .006}$ & $0.587^{\pm .005}$ & $3.599^{\pm .153}$ & $4.591^{\pm .026}$ & $9.473^{\pm .117}$ & $\mathbf{\textcolor{blue}{3.292^{\pm .081}}}$ \\
T2M \cite{t2m} & $0.361^{\pm .006}$ & $0.559^{\pm .007}$ & $0.681^{\pm .007}$ & $3.022^{\pm .107}$ & $3.488^{\pm .028}$ & $10.72^{\pm .145}$ & $2.052^{\pm .107}$ \\
\rowcolor{LightGray}
MDM$^{\S}$ \cite{MDM} & $0.164^{\pm .004}$ & $0.291^{\pm .004}$ & $0.396^{\pm .004}$ & $0.497^{\pm .021}$ & $9.191^{\pm .022}$ & $10.85^{\pm .109}$ & $1.907^{\pm .214}$ \\
\rowcolor{LightGray}
MotionDiffuse$^{\S}$ \cite{MotionDiffuse} & $0.417^{\pm .004}$ & $0.621^{\pm .004}$ & $0.739^{\pm .004}$ & $1.954^{\pm .064}$ & $2.958^{\pm .005}$ & $\mathbf{\textcolor{blue}{11.10^{\pm .143}}}$ & $0.730^{\pm .013}$ \\
\rowcolor{LightGray}
MLD$^{\S}$ \cite{MLD} & $0.390^{\pm .008}$ & $0.609^{\pm .008}$ & $0.734^{\pm .007}$ & $0.404^{\pm .027}$ & $3.204^{\pm .027}$ & $10.80^{\pm .117}$ & $2.192^{\pm .071}$ \\
\rowcolor{LightGray}
Fg-T2M$^{\S}$ \cite{Fg-T2M} & $0.418^{\pm .005}$ & $0.626^{\pm .004}$ & $0.745^{\pm .004}$ & $0.571^{\pm .047}$ & $3.114^{\pm .015}$ & $10.93^{\pm .083}$ & $1.019^{\pm .029}$ \\
\rowcolor{LightGray}
M2DM$^{\S}$ \cite{M2DM} & $0.416^{\pm .004}$ & $0.628^{\pm .004}$ & $0.743^{\pm .004}$ & $0.515^{\pm .029}$ & $3.015^{\pm .017}$ & $\mathbf{\textcolor{red}{11.417^{\pm .97}}}$ & $\mathbf{\textcolor{red}{3.325^{\pm .37}}}$ \\
T2M-GPT \cite{T2M-GPT} & $0.402^{\pm .006}$ & $0.619^{\pm .005}$ & $0.737^{\pm .006}$ & $0.717^{\pm .041}$ & $3.053^{\pm .026}$ & $10.86^{\pm .094}$ & $1.912^{\pm .036}$ \\
AttT2M \cite{AttT2M} & $0.413^{\pm .006}$ & $0.632^{\pm .006}$ & $0.751^{\pm .006}$ & $0.870^{\pm .039}$ & $3.039^{\pm .021}$ & $10.96^{\pm .123}$ & $2.281^{\pm .047}$ \\
\rowcolor{LightGray}
MMM$^{\S}$ \cite{MMM} & $0.404^{\pm .005}$ & $0.621^{\pm .005}$ & $0.744^{\pm .004}$ & $0.316^{\pm .028}$ & $2.977^{\pm .019}$ & $10.910^{\pm .101}$ & $1.232^{\pm .039}$ \\
\rowcolor{LightGray}
MoMask$^{\S}$ \cite{MoMask} & $0.433^{\pm .007}$ & $0.656^{\pm .005}$ & $0.781^{\pm .005}$ & $0.204^{\pm .011}$ & $2.779^{\pm .022}$ & - & $1.131^{\pm .043}$ \\

\toprule

BAMM (ours)  & $\textcolor{red}{\mathbf{0.438^{\pm .009}}}$ & $\textcolor{red}{\mathbf{0.661^{\pm .009}}}$ & $\textcolor{red}{\mathbf{0.788^{\pm .005}}}$ & $\textcolor{red}{\mathbf{0.183^{\pm .013}}}$ & $\textcolor{red}{\mathbf{2.723^{\pm .026}}}$ & $11.008^{\pm .094}$ & $1.609^{\pm .065}$ \\

\rowcolor{LightGray}
BAMM (ours)$^{\S}$  & $\textcolor{red}{\mathbf{0.436^{\pm .007}}}$ & $\textcolor{red}{\mathbf{0.660^{\pm .006}}}$ & $\textcolor{red}{\mathbf{0.791^{\pm .005}}}$ & $\textcolor{red}{\mathbf{0.200^{\pm .011}}}$ & $\textcolor{red}{\mathbf{2.714^{\pm .016}}}$ & $10.914^{\pm .097}$ & $1.517^{\pm .058}$ \\

\bottomrule
\end{tabular}
}
\label{tab:kit}
\vspace{-6pt}
\end{table*}

\textbf{Quantitative Results.} 
We evaluate our model on the HumanML3D \cite{t2m} and KIT-ML \cite{KIT} datasets, reporting the results in Table \ref{tab:humanml3d} and \ref{tab:kit}, respectively, in comparison to state-of-the-art methods. Following the standard evaluation protocol from \cite{t2m}, we report the average of 20 generations with a 95\% confidence interval. Our model consistently outperforms other methods in terms of R-precision, FID, and MM-Distance, while maintaining comparability in terms of Diversity and Multimodal Distance, which typically represent the trade-off between high quality and diversity. This suggests that our model generates very high-quality outputs while retaining a good degree of diversity. Moreover, whereas most models utilize the ground truth length for evaluation, our model supports both predicted length and takes length as an input, outperforming other methods in both scenarios. We further investigate the effect of a separate length estimator which can significantly impact performance in Section \ref{sec:predlen_edit}.

\begin{figure*}[ht]
 \vspace{-4mm}
 \centering
  \includegraphics[width=.8\textwidth]{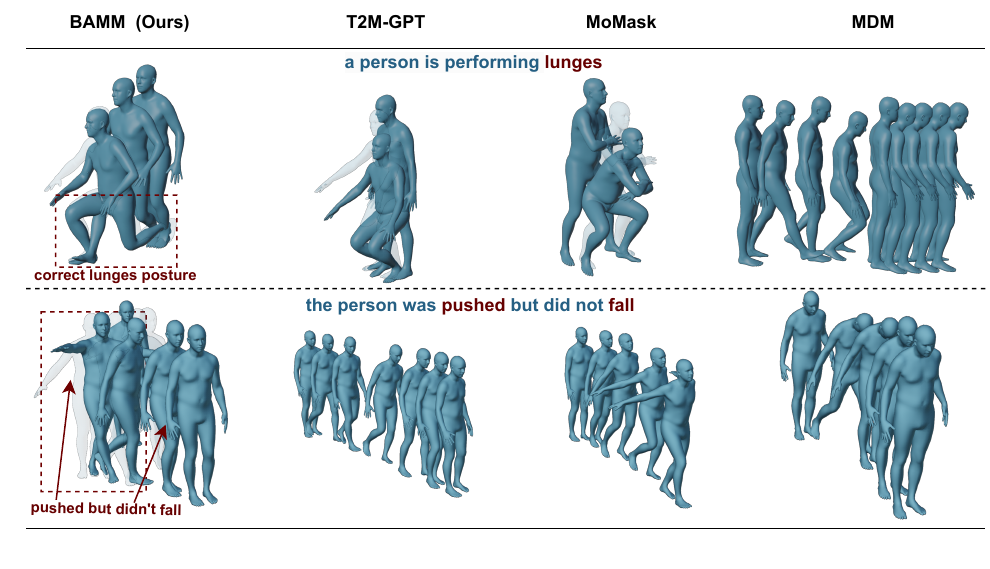}
  \vspace{-5mm}
  \caption{Visualization comparison of textual to motion to state-of-the-art methods. BAMM and T2M-GPT do not require motion length as an input. We use a pre-trained length estimator from \cite{t2m} for MoMask and MDM. BAMM generates higher quality and is more correlated with textual descriptions.}
  \label{fig:sota_comparison}
\end{figure*}

\begin{figure*}[ht]
 \centering
  \includegraphics[width=.8\textwidth]{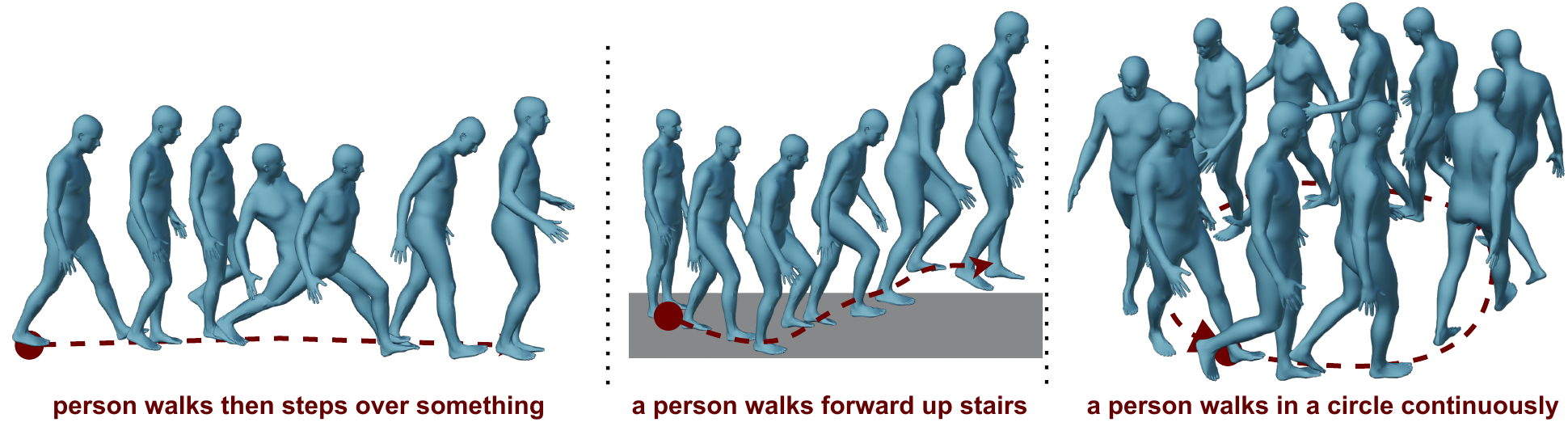}
  \caption{Visualization of text-to-motion generation by BAMM. BAMM can generate high-quality motion with complex descriptions, such as intricate trajectories and interactions with invisible objects.}
  \label{fig:text-to-motion}
    \vspace{-3mm}
\end{figure*}

\begin{figure*}[ht]
 \centering
  \includegraphics[width=1\textwidth]{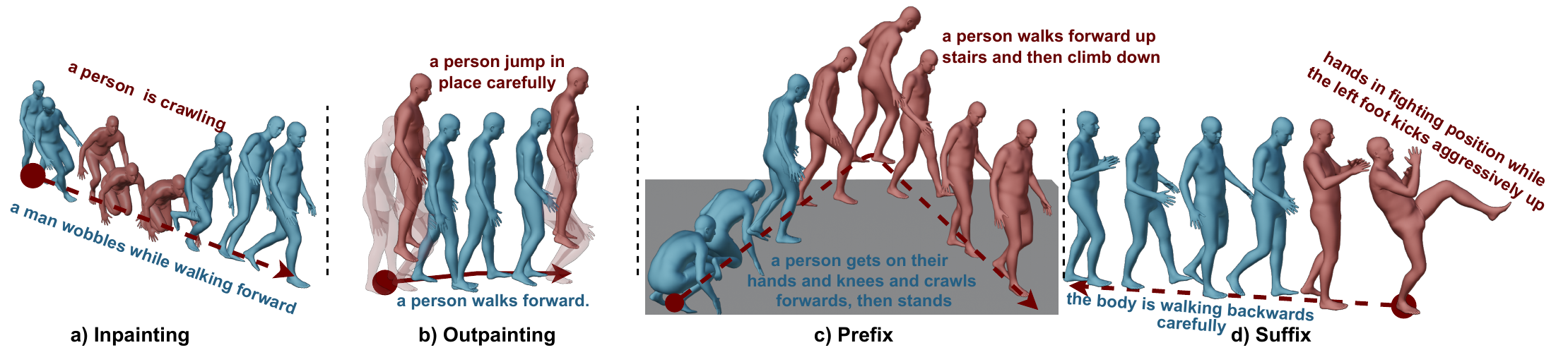}
  \caption{Visualization of temporal editing tasks, inpainting (in-betweening), outpainting, prefix, and suffix where \textcolor{MidnightBlue}{\textbf{blue}} indicates conditioned motion and \textcolor{Maroon}{  \textbf{red}} refers to generated parts.}
  \label{fig:editing_tasks}
\end{figure*}

\textbf{Qualitative Results.}
Fig. \ref{fig:sota_comparison} presents a qualitative comparison with T2M-GPT \cite{T2M-GPT}, MoMask \cite{MoMask}, and MDM \cite{MDM}. BAMM and T2M-GPT generate motion without requiring input length. Despite this, BAMM accurately generates motion from textual descriptions even without input length. We utilize a pre-trained length estimator from \cite{t2m} for MoMask \cite{MoMask} and MDM \cite{MDM}. Notably, BAMM generates motion accurately aligned with the provided text, whereas T2M-GPT and MoMask produce erroneous motion, and MDM generates entirely inaccurate motion. Additional visualizations for BAMM are shown in Fig. \ref{fig:text-to-motion}, indicating its ability to generate high detail such as complex trajectories and interactions with invisible objects. Moreover, Fig. \ref{fig:editing_tasks} demonstrates BAMM's capability in various temporal editing tasks, including inpainting, outpainting, prefix, and suffix. 

\subsection{Length Prediction and Editablity}
\label{sec:predlen_edit}
\textbf{Predict Length vs ground truth length.}
In real-world scenarios, the ground truth length is often unknown. Many methods cannot predict length autonomously. Consequently, they evaluate using ground truth length, indicated by \hl{gray highlight $^{\S}$} in Table \ref{tab:humanml3d} and Table \ref{tab:kit}. We illustrate the impact of inaccurate length prediction by comparing our model to state-of-the-art denoising models that require motion length as an input, such as MoMask and MMM. In particular, both models take the estimated length obtained from a standalone length predictor pretrained on the HumanML3D dataset \cite{t2m}. Table \ref{tab:predlen} reveals that the predicted length significantly worsens MoMask's FID score, from 0.045 to 0.090, while MMM's R-precision Top1 drops from 0.515 to 0.504. In contrast, our BAMM not only maintains the best performance but also outperforms MMM and MoMask in both scenarios.

\begin{table*}[ht]
\centering
\caption{\textbf{Comparison of text-conditional motion synthesis using predicted and ground truth length on HumanML3D\cite{t2m} dataset}}
\vspace{-7pt}
\scalebox{.6}{
\begin{tabular}{lcccccccc} 
\hline
\multirow{2}{*}{Length} & \multirow{2}{*}{Methods} & \multicolumn{3}{c}{R-Precision $\uparrow$} & \multirow{2}{*}{FID $\downarrow$} & \multirow{2}{*}{MM-Dist $\downarrow$} & \multirow{2}{*}{Diversity $\uparrow$} & \multirow{2}{*}{MModality $\uparrow$}        \\ 
\cline{3-5}
& & Top-1 $\uparrow$& Top-2 $\uparrow$& Top-3 $\uparrow$& & & & \\ 
\toprule
Ground       & MMM & 0.515 & 0.708 & 0.804 & 0.089 & 2.926 & 9.577 & 1.226 \\
 Truth      & MoMask & 0.521 & 0.713 & 0.807 & \textbf{0.045} & 2.958 & - & 1.241 \\
             & BAMM (Ours) & \textbf{0.522} & \textbf{0.715} & \textbf{0.808} & 0.055 & \textbf{2.936} & \textbf{9.636} & \textbf{1.732} \\
\toprule
         & MMM & 0.504 & 0.696 & 0.794 & 0.080 & 2.998 & 9.411 & 1.164 \\
Predicted & MoMask & 0.522 & 0.715 & 0.811 & 0.090 & 2.945 & 9.647 & 1.239 \\
         & BAMM (Ours) & \textbf{0.525} & \textbf{0.720} & \textbf{0.814} & \textbf{0.055} & \textbf{2.919} & \textbf{9.717} & \textbf{1.687} \\

\bottomrule
\end{tabular}
}
\label{tab:predlen}
\vspace{-6pt}
\end{table*}

\textbf{Editability.}
We conduct experiments in four temporal editing tasks, namely inpainting (in-betweening), outpainting, prefix, and suffix, comparing them with MDM \cite{MDM} and MoMask \cite{MoMask} on the HumanML3D dataset. Inpainting is evaluated by generating 50\% of the motion sequence given the first and last 25\%. Outpainting is its opposite counterpart. Prefix is conditioned by the first 50\% of the ground truth motion and generates the remaining portion. Similarly, suffix operates in the reverse manner. Note that editing tasks require prior knowledge of conditioned motion and motion length, for which MDM and MoMask are specifically designed. Despite this, Table \ref{tab:edit} shows that our model significantly outperforms MDM in all editing tasks and surpasses MoMask in outpainting, while performing equally well in the other tasks.

\begin{table*}[ht]
\centering
\vspace{-10pt}
\caption{\textbf{Evaluation on temporal editing tasks on HumanML3D \cite{t2m} dataset}}
\vspace{-5pt}
\scalebox{.6}{
\begin{tabular}{lcccccccc} 
\hline
\multirow{2}{*}{Tasks} &  \multirow{2}{*}{Methods} & \multicolumn{3}{c}{R-Precision $\uparrow$} & \multirow{2}{*}{FID $\downarrow$} & \multirow{2}{*}{MM-Dist $\downarrow$} & \multirow{2}{*}{Diversity $\uparrow$}        \\ 
\cline{3-5}
& & Top-1 $\uparrow$& Top-2 $\uparrow$& Top-3 $\uparrow$& & &  \\ 
\toprule
Temporal Inpainting                  & MDM    & 0.391          & 0.578          & 0.692          & 2.362          & 3.859          & 8.014          \\
(In-betweening)                    & MoMask & 0.534          & 0.727          & 0.82           & \textbf{0.04}  & 2.878          & \textbf{9.64}  \\
                                   & BAMM   & \textbf{0.535} & \textbf{0.729} & \textbf{0.821} & 0.056          & \textbf{2.863} & 9.629          \\
\toprule
Temporal Outpainting               & MDM    & 0.415          & 0.613          & 0.727          & 2.057          & 3.619          & 8.199          \\
                                   & MoMask & 0.531          & 0.726          & 0.818          & 0.057          & 2.889          & 9.619          \\
                                   & BAMM   & \textbf{0.535} & \textbf{0.73}  & \textbf{0.822} & \textbf{0.056} & \textbf{2.856} & \textbf{9.659} \\
\toprule
Prefix                             & MDM    & 0.42           & 0.613          & 0.725          & 1.46           & 3.563          & 8.312          \\
                                   & MoMask & \textbf{0.536} & \textbf{0.73}  & \textbf{0.822} & 0.06           & 2.875          & 9.607          \\
                                   & BAMM   & 0.532          & 0.727          & 0.821          & \textbf{0.058} & \textbf{2.868} & \textbf{9.612} \\
\toprule
Suffix                             & MDM    & 0.403          & 0.597          & 0.711          & 2.562          & 3.731          & 8.088          \\
                                   & MoMask & \textbf{0.532} & \textbf{0.726} & \textbf{0.819} & 0.052          & \textbf{2.881} & 9.659          \\
                                   & BAMM   & 0.527          & 0.72           & 0.814          & \textbf{0.05}  & 2.891          & \textbf{9.721} \\

\bottomrule
\end{tabular}
}
\label{tab:edit}
\vspace{-20pt}
\end{table*}

\section{Ablation Study}
\label{sec:ablation}



In the ablation study, we study the impacts of the adaptive classifier-free guidance scales in each iteration of the Masked Self-attention Transformer, the masking strategy, and the number of iterations, as shown in Table \ref{tab:ablation}.

\begin{table*}[ht]
\centering
\caption{\textbf{Ablation Study}}
\vspace{-10pt}
\scalebox{.6}{
\begin{tabular}{cccccccccc} 
\hline
\multirow{2}{*}{Ablations} & \multirow{2}{*}{Types} & &\multicolumn{3}{c}{R-Precision $\uparrow$} & \multirow{2}{*}{FID $\downarrow$} & \multirow{2}{*}{MM-Dist $\downarrow$} & \multirow{2}{*}{Diversity $\uparrow$} & \multirow{2}{*}{MModality $\uparrow$}        \\ 
\cline{4-6}
& & & Top-1 $\uparrow$& Top-2 $\uparrow$& Top-3 $\uparrow$& & & & \\ 

\toprule
              &  1st iter & 2nd iter &  &   &   &   &   &  &   \\
\cmidrule{2-3}
              & CFG=2           & CFG=3& 0.517  & 0.711 & 0.805 & 0.105  & 2.956 & 9.84 & \textbf{1.907} \\
  CFG of      & CFG=3           & CFG=3& 0.521  & 0.714 & 0.808 & 0.07 & 2.944 & \textbf{9.777} & 1.766 \\
1st iteration & \textbf{CFG=4}  & CFG=3& \textbf{0.525}  & \textbf{0.72}  & \textbf{0.814} & 0.055 & \textbf{2.919} & 9.717 & 1.687 \\
              & CFG=5           & CFG=3& 0.522  & 0.716 & 0.81  & \textbf{0.052} & 2.927 & 9.647 & 1.691 \\
              & CFG=6           & CFG=3& 0.52   & 0.713 & 0.81  & 0.06 & 2.94   & 9.621 & 1.697 \\
\cmidrule{2-3}
              & CFG=4           & CFG=1& 0.521  & 0.716 & 0.81  & 0.058 & 2.943 & 9.743 & \textbf{1.744} \\
CFG of        & CFG=4           & CFG=2& 0.524  & 0.719 & \textbf{0.814} & 0.057 & 2.924 & \textbf{9.752} & 1.698 \\
2nd iteration & CFG=4           & \textbf{CFG=3}& \textbf{0.525}  & \textbf{0.72}  & \textbf{0.814} & \textbf{0.055} & \textbf{2.919} & 9.717 & 1.687 \\
              & CFG=4           & CFG=4& 0.522  & 0.719 & 0.812 & 0.056 & 2.924 & 9.691 & 1.698 \\
              & CFG=4           & CFG=5& 0.521  & 0.717 & 0.812 & 0.065 & 2.931 & 9.638 & 1.735 \\
\toprule
              & 50\% of low confidence & & \textbf{0.525}   & 0.72  & 0.813 & 0.065 & 2.921 & 9.732 & 1.67  \\
  Mask         & confidence < .5         & & \textbf{0.525} & 0.718 & 0.81  & 0.064 & 2.923 & \textbf{9.765} & 1.656 \\
Strategy      & suffix                 & & 0.519  & 0.715 & 0.81  & 0.052 & 2.943 & 9.683  & \textbf{1.841} \\
              & \textbf{\%2=0}      & & \textbf{0.525}  & \textbf{0.72}  & \textbf{0.814} & \textbf{0.055} & \textbf{2.919} & 9.717 & 1.687 \\
\toprule
     \# of   &  1 iteration             & & 0.524 & 0.718 & 0.812 & 0.064 & 2.926 & 9.720 & \textbf{1.644} \\
iterations   &  \textbf{2 iterations}   & & \textbf{0.525} & \textbf{0.720} & \textbf{0.814} & \textbf{0.055} & 2.919 & 9.717 & 1.687 \\
            &  3 iterations             & & \textbf{0.525} & 0.719 & \textbf{0.814} & \textbf{0.055} & \textbf{2.917} & \textbf{9.727} & 1.69  \\


\bottomrule
\end{tabular}
}
\label{tab:ablation}
\vspace{-6pt}
\end{table*}

\textbf{Adaptive Classifier-Free Guidance Scales (CFG):} We conducted experiments with different CFG values for both the first and second iterations and found that CFG=4 in the first iteration and CFG=3 in the second iteration works best. Although CFG=5 in the first iteration yields a better FID score, the other scores are worse. \textbf{Mask Strategy:} The first strategy, {\it ``50\% of low confidence''}, applies mask attention on 50\% of the lowest confident positions from the first iteration. Similarly, {\it ``confidence < .5''} uses low confidence but sets a threshold, masking out positions where confidence is below 0.5. Specifically, the former strictly masks out 50\% of the total token sequence while the latter masks the token with a confidence lower than .5. {\it ``Suffix''} strategy masks out the first 50\% of the token sequence, utilizing the remaining 50\% as condition tokens. Lastly, {\it ``\%2=0''} strategy masks every other token. BAMM produces decent performances across the tested masking strategies while the simple masking of every other token works the best. \textbf{Number of Iterations:} We conducted experiments with one to three iterations. ``1 iteration'' refers to the first iteration, as described in \ref{sec:inference}, which utilizes a unidirectional causal mask. Similarly, ``2 iterations'' involves applying bidirectional causal mask to the Masked Self-attention Transformer to re-predict the tokens from the first iteration, as illustrated in \ref{fig:inference}. In ``3 iterations'', we repeat the second iteration (bidirectional causal mask) but apply masks to 1/3 of the sequence to re-predict the motion tokens instead. The experiments indicate that ``2 iterations'' clearly demonstrate improvement over one iteration, while ``3 iterations'' do not significantly improve upon ``2 iterations''. This suggests that ``2 iterations'' are sufficient.

\section{Conclusion}
 We introduce the Bidirectional Autoregressive Motion Model (BAMM), a novel framework for text-to-motion generation. BAMM combines a motion tokenizer, which encodes 3D human motion into discrete latent tokens, with a masked self-attention transformer that autoregressively predicts the masked tokens through a masked casual attention approach. BAMM integrates of generative masked and autoregressive modeling into an unified framework, This features allows it to understand complex motion relationships and precisely map text inputs to high-quality motion outputs with adaptively adjusted sequence lengths. Our extensive testing on HumanML3D and KIT-ML datasets confirms BAMM's superiority in both qualitative and quantitative evaluations over existing methods.

%
%
\bibliographystyle{splncs04}
\bibliography{bib_dir/intro,bib_dir/t2m,bib_dir/diffusion,bib_dir/other,bib_dir/token}

\begin{thebibliography}{10}
\providecommand{\url}[1]{\texttt{#1}}
\providecommand{\urlprefix}{URL }
\providecommand{\doi}[1]{https://doi.org/#1}

\bibitem{cmu-mocap}
Cmu graphics lab motion capture database, \url{http://mocap.cs.cmu.edu/}, accessed: 2022-11-11

\bibitem{Language2Pose}
Ahuja, C., Morency, L.P.: Language2pose: Natural language grounded pose forecasting. In: 2019 International Conference on 3D Vision (3DV). pp. 719--728 (2019). \doi{10.1109/3DV.2019.00084}

\bibitem{DiffusMask}
Austin, J., Johnson, D.D., Ho, J., Tarlow, D., Van Den~Berg, R.: Structured denoising diffusion models in discrete state-spaces. Advances in Neural Information Processing Systems  \textbf{34},  17981--17993 (2021)

\bibitem{Borsos2023SoundStormEP}
Borsos, Z., Sharifi, M., Vincent, D., Kharitonov, E., Zeghidour, N., Tagliasacchi, M.: Soundstorm: Efficient parallel audio generation. ArXiv  \textbf{abs/2305.09636} (2023), \url{https://api.semanticscholar.org/CorpusID:258715176}

\bibitem{GPT}
Brown, T.B., Mann, B., Ryder, N., Subbiah, M., Kaplan, J., Dhariwal, P., Neelakantan, A., Shyam, P., Sastry, G., Askell, A., Agarwal, S., Herbert-Voss, A., Krueger, G., Henighan, T.J., Child, R., Ramesh, A., Ziegler, D.M., Wu, J., Winter, C., Hesse, C., Chen, M., Sigler, E., Litwin, M., Gray, S., Chess, B., Clark, J., Berner, C., McCandlish, S., Radford, A., Sutskever, I., Amodei, D.: Language models are few-shot learners. ArXiv  \textbf{abs/2005.14165} (2020), \url{https://api.semanticscholar.org/CorpusID:218971783}

\bibitem{Muse}
Chang, H., Zhang, H., Barber, J., Maschinot, A., Lezama, J., Jiang, L., Yang, M., Murphy, K.P., Freeman, W.T., Rubinstein, M., Li, Y., Krishnan, D.: Muse: Text-to-image generation via masked generative transformers. ArXiv  \textbf{abs/2301.00704} (2023), \url{https://api.semanticscholar.org/CorpusID:255372955}

\bibitem{MaskGIT}
Chang, H., Zhang, H., Jiang, L., Liu, C., Freeman, W.T.: Maskgit: Masked generative image transformer. 2022 IEEE/CVF Conference on Computer Vision and Pattern Recognition (CVPR) pp. 11305--11315 (2022), \url{https://api.semanticscholar.org/CorpusID:246680316}

\bibitem{MLD}
Chen, X., Jiang, B., Liu, W., Huang, Z., Fu, B., Chen, T., Yu, J., Yu, G.: Executing your commands via motion diffusion in latent space. 2023 IEEE/CVF Conference on Computer Vision and Pattern Recognition (CVPR) pp. 18000--18010 (2022), \url{https://api.semanticscholar.org/CorpusID:254408910}

\bibitem{BERT}
Devlin, J., Chang, M.W., Lee, K., Toutanova, K.: Bert: Pre-training of deep bidirectional transformers for language understanding. In: North American Chapter of the Association for Computational Linguistics (2019), \url{https://api.semanticscholar.org/CorpusID:52967399}

\bibitem{CogView2FA}
Ding, M., Zheng, W., Hong, W., Tang, J.: Cogview2: Faster and better text-to-image generation via hierarchical transformers. ArXiv  \textbf{abs/2204.14217} (2022), \url{https://api.semanticscholar.org/CorpusID:248476190}

\bibitem{Du2021GLMGL}
Du, Z., Qian, Y., Liu, X., Ding, M., Qiu, J., Yang, Z., Tang, J.: Glm: General language model pretraining with autoregressive blank infilling. In: Annual Meeting of the Association for Computational Linguistics (2021), \url{https://api.semanticscholar.org/CorpusID:247519241}

\bibitem{mask-predict}
Ghazvininejad, M., Levy, O., Liu, Y., Zettlemoyer, L.: Mask-predict: Parallel decoding of conditional masked language models. In: Conference on Empirical Methods in Natural Language Processing (2019), \url{https://api.semanticscholar.org/CorpusID:202538740}

\bibitem{hier}
Ghosh, A., Cheema, N., Oguz, C., Theobalt, C., Slusallek, P.: Synthesis of compositional animations from textual descriptions. 2021 IEEE/CVF International Conference on Computer Vision (ICCV) pp. 1376--1386 (2021), \url{https://api.semanticscholar.org/CorpusID:232404671}

\bibitem{MoMask}
Guo, C., Mu, Y., Javed, M.G., Wang, S., Cheng, L.: Momask: Generative masked modeling of 3d human motions (2023)

\bibitem{TM2T}
Guo, C., Xuo, X., Wang, S., Cheng, L.: Tm2t: Stochastic and tokenized modeling for the reciprocal generation of 3d human motions and texts. ArXiv  \textbf{abs/2207.01696} (2022), \url{https://api.semanticscholar.org/CorpusID:250280248}

\bibitem{t2m}
Guo, C., Zou, S., Zuo, X., Wang, S., Ji, W., Li, X., Cheng, L.: Generating diverse and natural 3d human motions from text. In: 2022 IEEE/CVF Conference on Computer Vision and Pattern Recognition (CVPR). pp. 5142--5151 (2022). \doi{10.1109/CVPR52688.2022.00509}

\bibitem{HumanAct12}
Guo, C., Zuo, X., Wang, S., Zou, S., Sun, Q., Deng, A., Gong, M., Cheng, L.: Action2motion: Conditioned generation of 3d human motions. Proceedings of the 28th ACM International Conference on Multimedia  (2020), \url{https://api.semanticscholar.org/CorpusID:220870974}

\bibitem{ImagenVideo}
Ho, J., Chan, W., Saharia, C., Whang, J., Gao, R., Gritsenko, A.A., Kingma, D.P., Poole, B., Norouzi, M., Fleet, D.J., Salimans, T.: Imagen video: High definition video generation with diffusion models. ArXiv  \textbf{abs/2210.02303} (2022), \url{https://api.semanticscholar.org/CorpusID:252715883}

\bibitem{DDPM}
Ho, J., Jain, A., Abbeel, P.: Denoising diffusion probabilistic models. ArXiv  \textbf{abs/2006.11239} (2020), \url{https://api.semanticscholar.org/CorpusID:219955663}

\bibitem{cfg}
Ho, J., Salimans, T.: Classifier-free diffusion guidance. arXiv preprint arXiv:2207.12598  (2022)

\bibitem{MotionGPT}
Jiang, B., Chen, X., Liu, W., Yu, J., Yu, G., Chen, T.: Motiongpt: Human motion as a foreign language. ArXiv  \textbf{abs/2306.14795} (2023), \url{https://api.semanticscholar.org/CorpusID:259262201}

\bibitem{FLAME}
Kim, J., Kim, J., Choi, S.: Flame: Free-form language-based motion synthesis \& editing. In: AAAI Conference on Artificial Intelligence (2022), \url{https://api.semanticscholar.org/CorpusID:251979380}

\bibitem{M2DM}
Kong, H., Gong, K., Lian, D., Mi, M.B., Wang, X.: Priority-centric human motion generation in discrete latent space. ArXiv  \textbf{abs/2308.14480} (2023), \url{https://api.semanticscholar.org/CorpusID:261245369}

\bibitem{AMASS}
Mahmood, N., Ghorbani, N., Troje, N.F., Pons-Moll, G., Black, M.J.: Amass: Archive of motion capture as surface shapes. 2019 IEEE/CVF International Conference on Computer Vision (ICCV) pp. 5441--5450 (2019), \url{https://api.semanticscholar.org/CorpusID:102351100}

\bibitem{GLIDE}
Nichol, A., Dhariwal, P., Ramesh, A., Shyam, P., Mishkin, P., McGrew, B., Sutskever, I., Chen, M.: Glide: Towards photorealistic image generation and editing with text-guided diffusion models. In: International Conference on Machine Learning (2021), \url{https://api.semanticscholar.org/CorpusID:245335086}

\bibitem{vqvae}
van~den Oord, A., Vinyals, O., Kavukcuoglu, K.: Neural discrete representation learning. ArXiv  \textbf{abs/1711.00937} (2017), \url{https://api.semanticscholar.org/CorpusID:20282961}

\bibitem{TEMOS}
Petrovich, M., Black, M.J., Varol, G.: Temos: Generating diverse human motions from textual descriptions. ArXiv  \textbf{abs/2204.14109} (2022), \url{https://api.semanticscholar.org/CorpusID:248476220}

\bibitem{TMR}
Petrovich, M., Black, M.J., Varol, G.: Tmr: Text-to-motion retrieval using contrastive 3d human motion synthesis. ArXiv  \textbf{abs/2305.00976} (2023), \url{https://api.semanticscholar.org/CorpusID:258436810}

\bibitem{MMM}
Pinyoanuntapong, E., Wang, P., Lee, M., Chen, C.: Mmm: Generative masked motion model. In: Proceedings of the IEEE/CVF Conference on Computer Vision and Pattern Recognition (CVPR) (2024)

\bibitem{KIT}
Plappert, M., Mandery, C., Asfour, T.: The {KIT} motion-language dataset. Big Data  \textbf{4}(4),  236--252 (dec 2016). \doi{10.1089/big.2016.0028}, \url{http://dx.doi.org/10.1089/big.2016.0028}

\bibitem{GlancingTF}
Qian, L., Zhou, H., Bao, Y., Wang, M., Qiu, L., Zhang, W., Yu, Y., Li, L.: Glancing transformer for non-autoregressive neural machine translation. In: Annual Meeting of the Association for Computational Linguistics (2020), \url{https://api.semanticscholar.org/CorpusID:221150562}

\bibitem{CLIP}
Radford, A., Kim, J.W., Hallacy, C., Ramesh, A., Goh, G., Agarwal, S., Sastry, G., Askell, A., Mishkin, P., Clark, J., Krueger, G., Sutskever, I.: Learning transferable visual models from natural language supervision. In: International Conference on Machine Learning (2021), \url{https://api.semanticscholar.org/CorpusID:231591445}

\bibitem{Imagen}
Saharia, C., Chan, W., Saxena, S., Li, L., Whang, J., Denton, E.L., Ghasemipour, S.K.S., Ayan, B.K., Mahdavi, S.S., Lopes, R.G., Salimans, T., Ho, J., Fleet, D.J., Norouzi, M.: Photorealistic text-to-image diffusion models with deep language understanding. ArXiv  \textbf{abs/2205.11487} (2022), \url{https://api.semanticscholar.org/CorpusID:248986576}

\bibitem{Make-A-Video}
Singer, U., Polyak, A., Hayes, T., Yin, X., An, J., Zhang, S., Hu, Q., Yang, H., Ashual, O., Gafni, O., Parikh, D., Gupta, S., Taigman, Y.: Make-a-video: Text-to-video generation without text-video data. ArXiv  \textbf{abs/2209.14792} (2022), \url{https://api.semanticscholar.org/CorpusID:252595919}

\bibitem{DDIM}
Song, J., Meng, C., Ermon, S.: Denoising diffusion implicit models. ArXiv  \textbf{abs/2010.02502} (2020), \url{https://api.semanticscholar.org/CorpusID:222140788}

\bibitem{MotionCLIP}
Tevet, G., Gordon, B., Hertz, A., Bermano, A.H., Cohen-Or, D.: Motionclip: Exposing human motion generation to clip space. In: European Conference on Computer Vision (2022), \url{https://api.semanticscholar.org/CorpusID:247450907}

\bibitem{MDM}
Tevet, G., Raab, S., Gordon, B., Shafir, Y., Cohen-Or, D., Bermano, A.H.: Human motion diffusion model. ArXiv  \textbf{abs/2209.14916} (2022), \url{https://api.semanticscholar.org/CorpusID:252595883}

\bibitem{Vaswani2017AttentionIA}
Vaswani, A., Shazeer, N.M., Parmar, N., Uszkoreit, J., Jones, L., Gomez, A.N., Kaiser, L., Polosukhin, I.: Attention is all you need. In: Neural Information Processing Systems (2017), \url{https://api.semanticscholar.org/CorpusID:13756489}

\bibitem{Phenaki}
Villegas, R., Babaeizadeh, M., Kindermans, P.J., Moraldo, H., Zhang, H., Saffar, M.T., Castro, S., Kunze, J., Erhan, D.: Phenaki: Variable length video generation from open domain textual description. ArXiv  \textbf{abs/2210.02399} (2022), \url{https://api.semanticscholar.org/CorpusID:252715594}

\bibitem{Wang2023NeuralCL}
Wang, C., Chen, S., Wu, Y., Zhang, Z.H., Zhou, L., Liu, S., Chen, Z., Liu, Y., Wang, H., Li, J., He, L., Zhao, S., Wei, F.: Neural codec language models are zero-shot text to speech synthesizers. ArXiv  \textbf{abs/2301.02111} (2023), \url{https://api.semanticscholar.org/CorpusID:255440307}

\bibitem{Fg-T2M}
Wang, Y., Leng, Z., Li, F.W.B., Wu, S.C., Liang, X.: Fg-t2m: Fine-grained text-driven human motion generation via diffusion model. ArXiv  \textbf{abs/2309.06284} (2023), \url{https://api.semanticscholar.org/CorpusID:261697123}

\bibitem{CrossModalRF}
Yan, S., Liu, Y., Wang, H., Du, X., Liu, M., Liu, H.: Cross-modal retrieval for motion and text via doptriple loss (2023), \url{https://api.semanticscholar.org/CorpusID:263610212}

\bibitem{Parti}
Yu, J., Xu, Y., Koh, J.Y., Luong, T., Baid, G., Wang, Z., Vasudevan, V., Ku, A., Yang, Y., Ayan, B.K., Hutchinson, B.C., Han, W., Parekh, Z., Li, X., Zhang, H., Baldridge, J., Wu, Y.: Scaling autoregressive models for content-rich text-to-image generation. Trans. Mach. Learn. Res.  \textbf{2022} (2022), \url{https://api.semanticscholar.org/CorpusID:249926846}

\bibitem{Zeghidour2021SoundStreamAE}
Zeghidour, N., Luebs, A., Omran, A., Skoglund, J., Tagliasacchi, M.: Soundstream: An end-to-end neural audio codec. IEEE/ACM Transactions on Audio, Speech, and Language Processing  \textbf{30},  495--507 (2021), \url{https://api.semanticscholar.org/CorpusID:236149944}

\bibitem{T2M-GPT}
Zhang, J., Zhang, Y., Cun, X., Huang, S., Zhang, Y., Zhao, H., Lu, H., Shen, X.: Generating human motion from textual descriptions with discrete representations. 2023 IEEE/CVF Conference on Computer Vision and Pattern Recognition (CVPR) pp. 14730--14740 (2023), \url{https://api.semanticscholar.org/CorpusID:255942203}

\bibitem{MotionDiffuse}
Zhang, M., Cai, Z., Pan, L., Hong, F., Guo, X., Yang, L., Liu, Z.: Motiondiffuse: Text-driven human motion generation with diffusion model. ArXiv  \textbf{abs/2208.15001} (2022), \url{https://api.semanticscholar.org/CorpusID:251953565}

\bibitem{M6UFCUM}
Zhang, Z., Ma, J., Zhou, C., Men, R., Li, Z., Ding, M., Tang, J., Zhou, J., Yang, H.: M6-ufc: Unifying multi-modal controls for conditional image synthesis via non-autoregressive generative transformers (2021), \url{https://api.semanticscholar.org/CorpusID:237204528}

\bibitem{UFCBERTUM}
Zhang, Z., Ma, J., Zhou, C., Men, R., Li, Z., Ding, M., Tang, J., Zhou, J., Yang, H.: Ufc-bert: Unifying multi-modal controls for conditional image synthesis. In: Neural Information Processing Systems (2021), \url{https://api.semanticscholar.org/CorpusID:235253928}

\bibitem{AttT2M}
Zhong, C., Hu, L., Zhang, Z., Xia, S.: Attt2m: Text-driven human motion generation with multi-perspective attention mechanism. ArXiv  \textbf{abs/2309.00796} (2023), \url{https://api.semanticscholar.org/CorpusID:261530775}

\end{thebibliography}
\clearpage
\setcounter{page}{1}
\onecolumn
\appendix

\begin{center} 
    \centering
    \textbf{\large BAMM: Bidirectional Autoregressive Motion Model}
\end{center}
\begin{center} 
    \centering
    \large Supplementary Material
\end{center}

\section{Overview}
\label{sec:Summary}
The supplementary material is organized into the following sections:
\begin{itemize}
    \item Section \ref{sec:lengpred_lengthrestriction}: Length prediction vs length restriction
    \item Section \ref{sec:len_div_high_quality}: Length diversity with high-quality motion generation.
    \item Section \ref{sec:temporal_motion_editing}: Temporal Motion Editing
    \item Section \ref{sec:implementation_details}: Implementation Details
    \item Section \ref{sec:limitation}: Limitation
\end{itemize}



\section{ Length prediction vs length restriction} 
\label{sec:lengpred_lengthrestriction}

 \begin{figure*}[ht]
 \centering
  \includegraphics[width=.5\textwidth]{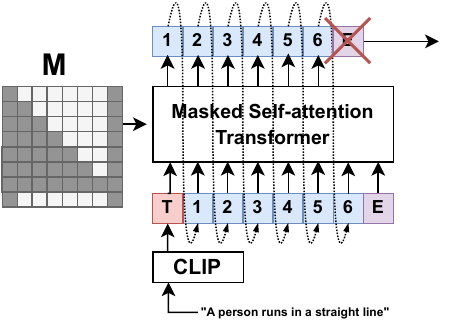}
  \caption{Generate motion with length constrain by input \texttt{[END]} as a condition and remove \texttt{[END]} output prediction.}
  \label{fig:input_len}
\end{figure*}
\textbf{Length prediction.} Naturally, BAMM has the ability to predict the \texttt{[END]} token to stop generating when it seems appropriate which automatically predicts the correlated motion length from previous token conditions without relying on an external length estimator as shown in the first iteration of Fig. \ref{fig:inference}. 

\textbf{Length restriction.} In tasks such as temporal motion editing that require specific motion lengths, our model can generate motion constrained by input motion length. This is achieved by applying the \texttt{[END]} token as an input condition to constrain where generation should stop. During training, the \texttt{[END]} token is already randomly conditioned. However, in this scenario, \texttt{[END]} serves as an input condition rather than an output prediction. To ensure uninterrupted generation until reaching the desired length without prematurely stopping due to \texttt{[END]} predictions, we force the model to predict only the $K$ indices of the codebook, explicitly excluding \texttt{[END]} predictions from the output logits. Therefore, the first iteration in Fig. \ref{fig:inference} can be modified to Fig. \ref{fig:input_len}.


\section{Length diversity with high-quality motion generation}
\label{sec:len_div_high_quality}
\begin{figure*}[ht]
 \centering
  \includegraphics[width=1\textwidth]{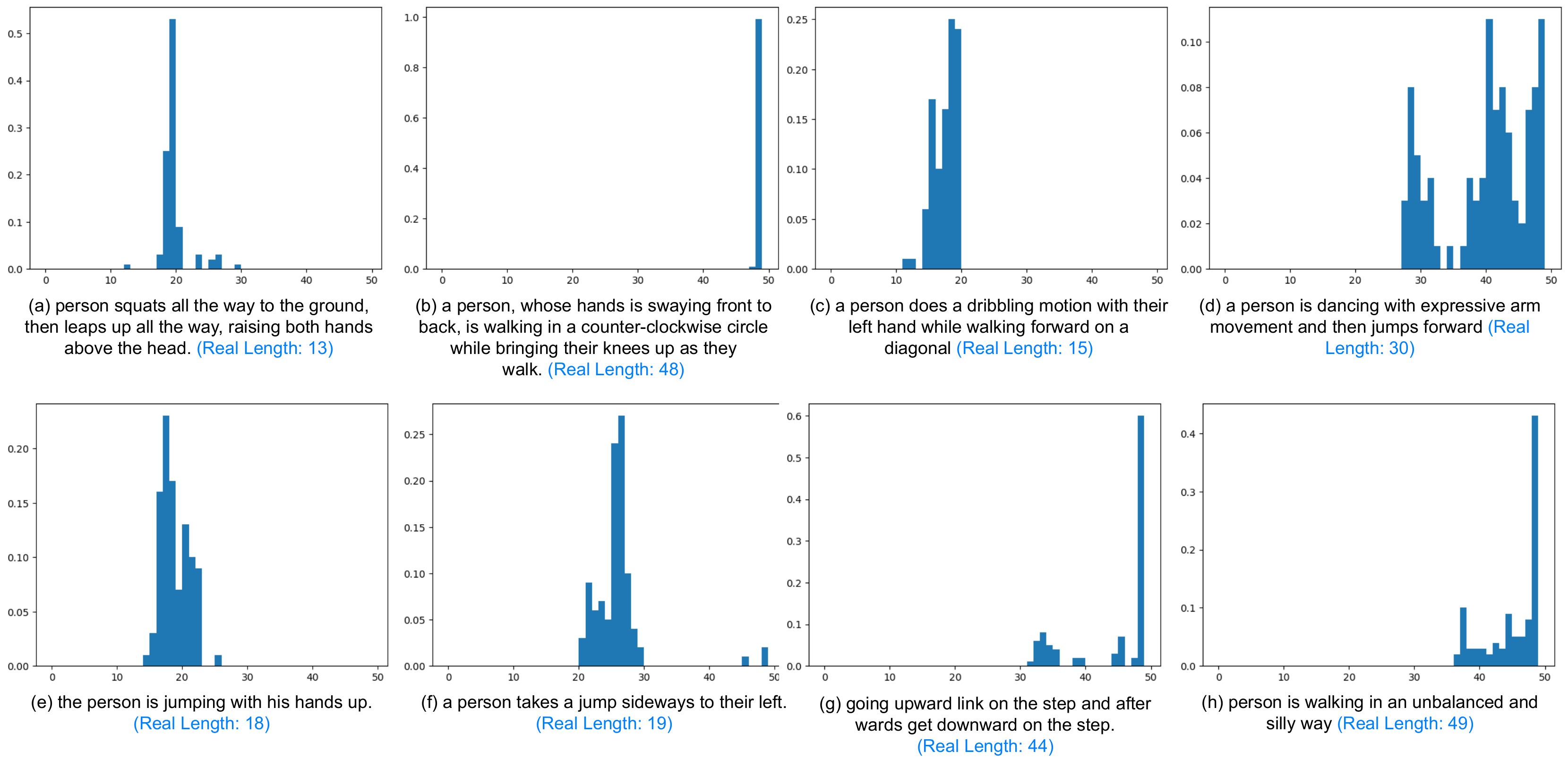}
  \caption{Histogram of motion token lengths. 1000 motions are generated for each textual description to calculate the estimated probability density of the token length. The corresponding lengths from the dataset HumanML3D \cite{t2m} are called Real Length and highlighted in \textcolor{blue}{\textbf{blue} text}. The length of motion is four times the token length.
 }
  \label{fig:lengt_predictor_histogram}
\end{figure*}

The benefit of our BAMM model's integrated length predictor is that it enhances motion realism and quality, as the model can re-evaluate every iteration. Additionally, the generated length is of a broader range, reflecting the diversity of motion while being correlated with the currently generated motions. The histogram in Fig. \ref{fig:lengt_predictor_histogram} illustrates various motion token lengths generated from the same textual description. For each textual description example, we generate 1000 samples and calculate the probability density of the predicted number of token lengths. Given a prompt, the predicted length of BAMM is generally diverse. The motions involving detailed, lengthy and sequential actions tend to have the maximum motion length, aligning closely with the ground truth, as shown in Fig. \ref{fig:lengt_predictor_histogram} (b).

In contrast, the models that rely on separated length estimators not only suffer from inaccurate motion length, but also lack diversity in the generated motions. We demonstrate this effect on the experiment with a pre-trained length estimator from \cite{t2m} for MMM \cite{MMM} and MoMask \cite{MoMask}, both of which require input length methods. To investigate the impact of length diversity on the quality of generated motion, we compare the motion generation performance under two motion length sampling strategies: top-1 sampling and multinomial sampling In Table \ref{tab:predlen}. The top-1 sampling always chooses the motion length predicted with the highest probability or confidence by the length estimator.  Multinomial sampling generates random motion length drawn from the prediction probability or confidence distribution. As shown in Fig. \ref{fig:lengt_predictor_histogram}, both MMM and MoMask experience degraded performance in terms of R-precision and FID when multinomial sampling is adopted. This is because multinomial sampling can generate diverse motion lengths that the models cannot adapt to. 


\begin{table*}[ht]
\centering
\caption{\textbf{Comparison of text-conditional motion synthesis using different length samping stategies on HumanML3D\cite{t2m} dataset}}
\vspace{-7pt}
\scalebox{.90}{
\begin{tabular}{lcccccccc} 
\hline
\multirow{2}{*}{Methods} & \multicolumn{3}{c}{R-Precision $\uparrow$} & \multirow{2}{*}{FID $\downarrow$} & \multirow{2}{*}{MM-Dist $\downarrow$} & \multirow{2}{*}{Diversity $\uparrow$} & \multirow{2}{*}{MModality $\uparrow$}        \\ 
\cline{2-4}
&  Top-1 $\uparrow$& Top-2 $\uparrow$& Top-3 $\uparrow$& & & & \\ 
\toprule
MMM Top-1 & 0.504 & 0.696 & 0.794 & 0.080 & 2.998 & 9.411 & 1.164 \\
MMM Multinomial & 0.492 & 0.685 & 0.782 & 0.099 & 3.063 & 9.319 & 1.18 \\
\toprule
MoMask Top-1 & 0.522 & 0.715 & 0.811 & 0.090 & 2.945 & 9.647 & 1.239 \\
MoMask Multinomial & 0.520 & 0.713 & 0.809 & 0.120 & 2.957 & \textbf{9.731} & 1.235 \\
\toprule
BAMM (Ours) & \textbf{0.525} & \textbf{0.720} & \textbf{0.814} & \textbf{0.055} & \textbf{2.919} & 9.717 & \textbf{1.687} \\

\bottomrule
\end{tabular}
}
\label{tab:predlen_random}
\vspace{-6pt}
\end{table*}

\begin{figure*}[ht]
 \centering
  \includegraphics[width=1\textwidth]{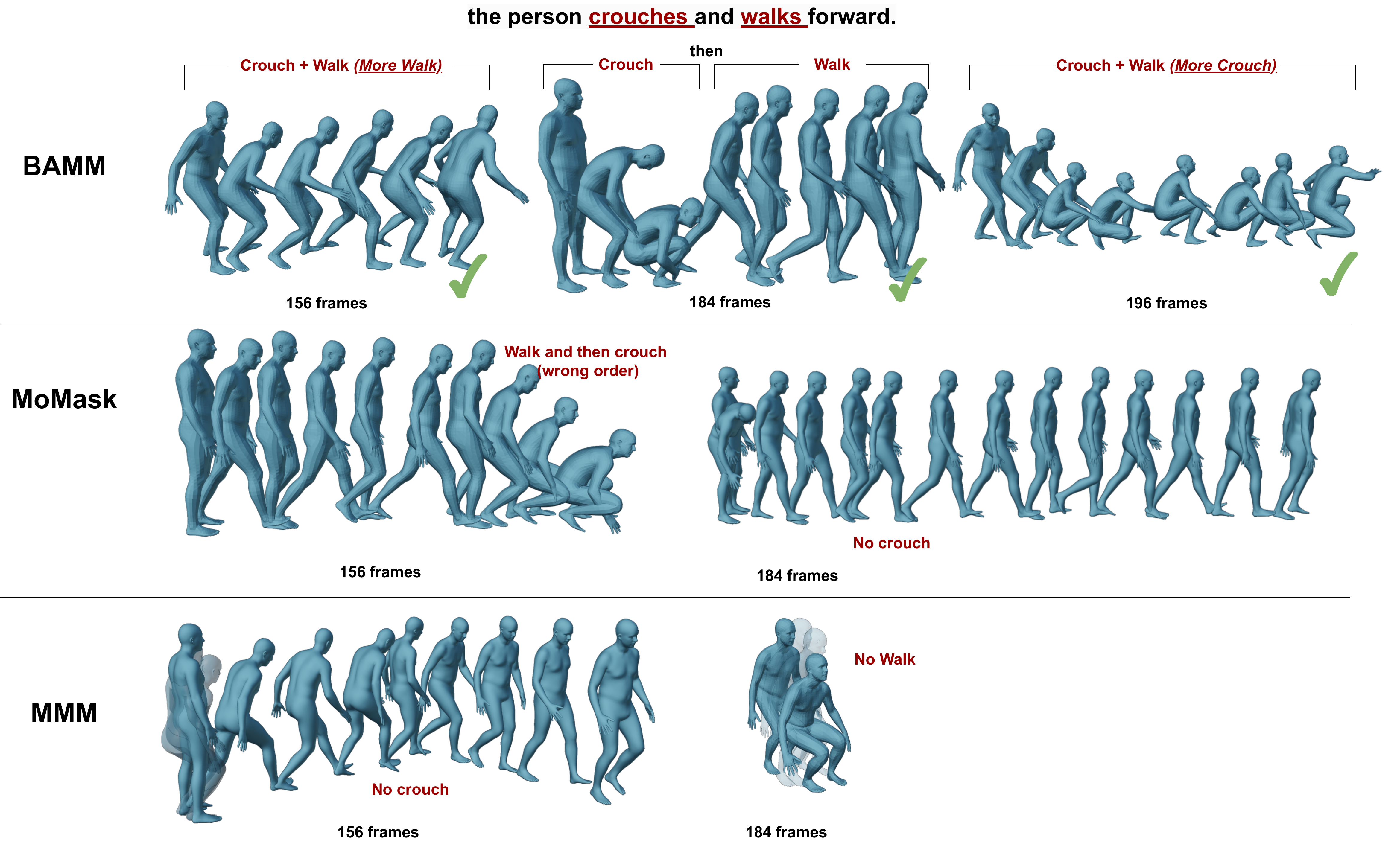}
  \caption{Visualization comparing different input lengths to state-of-the-art methods with the prompt "the person crouches and walks forward." with ground truth length of 196 frames. BAMM generates diverse motions correlated with various lengths while MMM and MoMask are sensitive to the different length inputs.
 }
  \label{fig:len_div}
\end{figure*}

In Fig. \ref{fig:len_div}, we demonstrate how a single prompt can lead to variations in motion, showcasing the diversity of motion correlated with different lengths. Using the textual description "the person crouches and walks forward.", we observe different interpretations of the motion generated by BAMM. For instance, the first and last samples show variations such as 'crouching while walking forward,' with the last sample exhibiting a deeper crouch. In contrast, the middle sample depicts separate actions of 'crouching' and 'walking forward.' Each sample has a unique length corresponding to its motion. However, MoMask and MMM are sensitive to varying lengths, resulting in inaccuracies in their generated motions when the lengths are not precise.


\section{Temporal Motion Editing}
\label{sec:temporal_motion_editing}
\begin{figure*}[ht]
 \centering
  \includegraphics[width=1\textwidth]{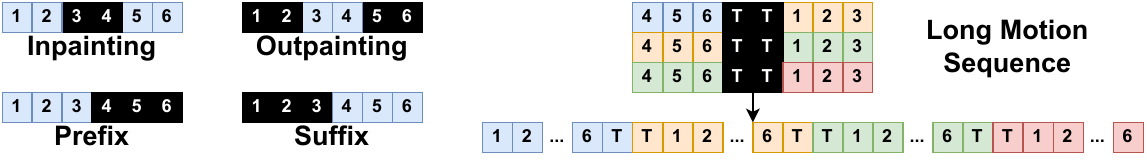}
  \caption{ Visualization of masking and conditional tokens for five temporal motion editing tasks: inpainting (in-betweening), outpainting, prefix, suffix, and long motion sequence. $\blacksquare$ indicates masked positions/areas 
 }
  \label{fig:edit_how_to}
\end{figure*}
Since our BAMM model can utilize conditional tokens as inputs for generation, temporal motion editing can be accomplished by predicting the tokens in the masked positions that need modifications, conditioned on the unmasked tokens and text prompt, as illustrated in Fig. \ref{fig:edit_how_to}. The visualization results are in Fig. \ref{fig:editing_tasks}. In addition, the editing tasks are performed in the zero-shot manner. This means that during the model training, we do not apply any specific masks that correspond to editing tasks as shown in Fig \ref{fig:edit_how_to} (left). Instead, we just randomly put $50\% - 100\%$ motion tokens in the masked areas. 

\begin{figure*}[ht]
 \centering
  \includegraphics[width=1\textwidth]{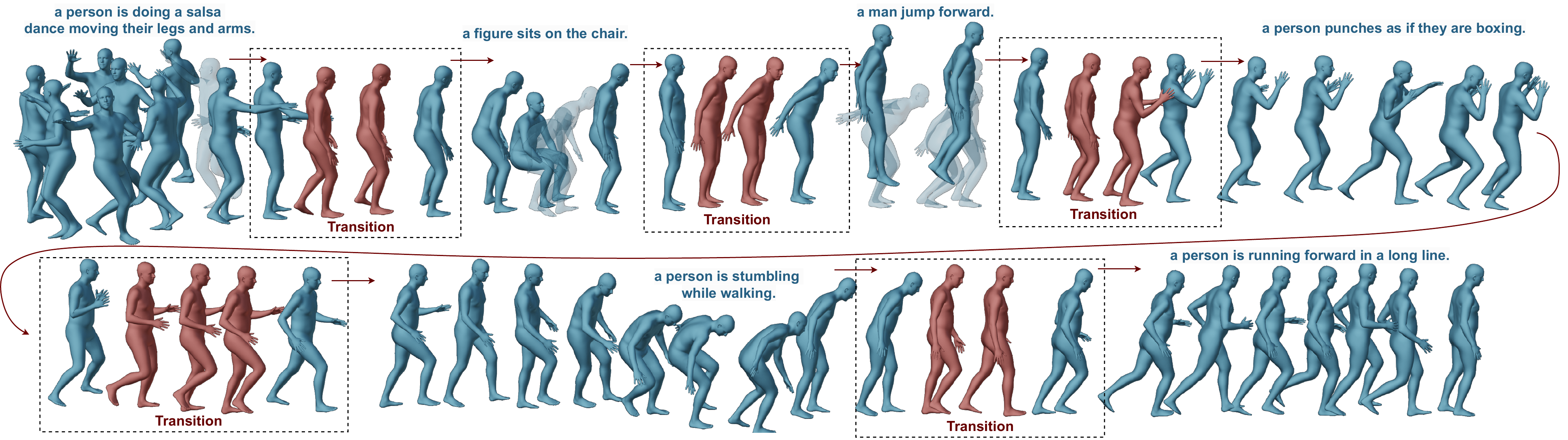}
  \caption{ Visualization of Long Motion Sequence where \textcolor{MidnightBlue}{\textbf{blue}} frames represent individual motion segments prompted by textual descriptions. \textcolor{Maroon}{\textbf{Red}} frames depict the intermediate transitions between these prompted segments, ensuring temporal coherence across the entire sequence. }
  \label{fig:bamm_longrange}
\end{figure*}

\textbf{Long Motion Sequence.} Generating arbitrarily long motions presents a challenge due to the limited length of motion data in available datasets such as HumanML3D \cite{t2m} and KIT \cite{KIT}, where no sample exceeds a duration of 10 seconds. To tackle this issue, we utilize the trained masked motion model as a prior for synthesizing long motion sequences without requiring additional training. Specifically, given a story consisting of multiple text prompts, our model first generates the motion token sequence for each prompt. Then, it generates transition motion tokens conditioned on the end of the previous motion sequence and the start of the next motion sequence.

\section{Implementation Details}
\label{sec:implementation_details}
The Motion tokenizer comprises six quantization layers, each with 512 codes and 512 embedding dimensions, along with skip connection and a dropout ratio of 0.2. Both the Masked Self-attention Transformer and Refinement Transformer consist of a six-layer encoder-only transformer architecture with six heads and an embedding size of 384. The batch size is set to 512 for both Motion Tokenizer and Masked Self-attention, while it is 64 for the Refinement Transformer. We use AdamW for optimization with a learning rate of 2e-4 which decreases by a factor of ten at 50,000 and 80,000 iterations. A masking ratio of 0.5 is applied for $\lambda$. During training, ground truth input is randomly replaced with random tokens with a probability of $\tau=0.5$. For HumanML3D, the CFG scales are set to 4, 3, and 6 for the first, the second stages, and Residual Motion Refinement, respectively. For KIT, the corresponding scales are 2, 2, and 6.

\section{Limitation}
\label{sec:limitation}
While BAMM offers high-quality motion generation, it is important to note that its processing speed is slower in comparison to parallel decoding methods like MMM or MoMask. This delay stems from BAMM's cascaded generation process, which includes an unidirectional autoregressive decoding process followed by a bidirectional autoregressive decoding procedure and a residual motion refinement step. Despite this, it is worth mentioning that BAMM still outperforms motion space diffusion techniques such as MDM and MotionDiffuse in terms of speed by a large margin. Additionally, with an average generation time of 0.411 seconds per sample on an NVIDIA RTX A5000, BAMM remains sufficiently fast for practical use.

\end{document}